%% file: root.tex
\documentclass[letterpaper,10pt,conference]{ieeeconf}
\usepackage{amssymb}
\usepackage{bbm}
\usepackage{amsfonts}
\usepackage{amsmath}
\usepackage[linesnumbered,ruled,vlined]{algorithm2e}
\usepackage{subcaption}
\usepackage{graphicx}
\usepackage{multicol}
\usepackage{multirow}
\usepackage{booktabs}
\usepackage{hyperref}
\usepackage{amssymb}
\usepackage{svg}
\input{newcommands}
\IEEEoverridecommandlockouts
\begin{document}

\title{Learning Neural Control Barrier Functions from Expert Demonstrations using Inverse Constraint Learning}
\author{
	Yuxuan Yang and Hussein Sibai \\
	Computer Science and Engineering Department\\
	Washington University in St. Louis \\
	\texttt{\{y.yuxuan,sibai\}@wustl.edu} \\
    % \vspace{0pt}\\
	% Hussein Sibai \\
	% Computer Science and Engineering Department \\
	% Washington University in St. Louis \\
	% \texttt{sibai@wustl.edu} \\
}
\maketitle

\begin{abstract}
    Safety is a fundamental requirement for autonomous systems operating in critical domains. Control barrier functions (CBFs) have been used to design safety filters that minimally alter nominal controls for such systems to maintain their safety. Learning neural CBFs has been proposed as a data-driven alternative for their computationally expensive optimization-based synthesis. However, it is often the case that the failure set of states that should be avoided is non-obvious or hard to specify formally, e.g., tailgating in autonomous driving, while a set of expert demonstrations that achieve the task and avoid the failure set is easier to generate. We use ICL to train a constraint function that classifies the states of the system under consideration to safe, i.e., belong to a controlled forward invariant set that is disjoint from the unspecified failure set, and unsafe ones, i.e., belong to the complement of that set. We then use that function to label a new set of simulated trajectories to train our neural CBF. We empirically evaluate our approach in four different environments, demonstrating that it outperforms existing baselines and achieves comparable performance to a neural CBF trained with the same data but annotated with ground-truth safety labels.
\end{abstract}

\section{Introduction}

Designing safe control policies for autonomous systems has been an ongoing challenge that limits their deployment in critical settings~\cite{learning_certified_control_chuchu_2023,CBF_survey_2019,safety_filter_survey_fisac_2024}. One approach for maintaining safety is the design of safety filters which adjust safety-agnostic nominal controls online when they are potentially safety-violating~\cite{cbf}. Control Barrier Functions (CBFs) define sets of states that can be kept invariant by choosing controls that satisfy linear inequalities, allowing the design for efficient safety filters in the form of quadratic programs~\cite{cbf,CBF_survey_2019}. They have been shown to be useful in maintaining safety in a variety of robotic applications, including bipedal robotic walking~\cite{CBF_bipedal_walking_ames_2015}, adaptive cruise control~\cite{CBF_cruise_control_ames_2014}, and space shuttles' docking~\cite{CBF_docking_panagou_2021}.  

%which in the presence of a reference controller controlled   \yuxuan{demonstrated} their effectiveness in \yuxuan{ensuring} the safety of control systems \cite{CBF_survey_2019}. 
When a set of {\em failure} (or {\em avoid}) states and the dynamics are known, CBFs can be synthesized using techniques such as Sum-of-Squares (SoS) \cite{Verification_and_Synthesis_using_SoS_Andrew_Clark_CDC_2021, Permissive_CBF_SOS_2018_Magnus} and Hamilton-Jacobi reachability analysis \cite{refining_CBF_using_hamilton_jacobi_IROS_2022,robust_CBF_for_safety_critical_control_Sylvia_Herbert_CDC_2021}. 
However, such correct-by-construction methods suffer from the curse-of-dimensionality. Supervised deep learning approaches have been proposed to train neural CBFs, which are easier to design and more scalable, at the expense of losing correctness guarantees~\cite{learning_certified_control_chuchu_2023,barriernet,sablas,vluefunctioniscbf}. Such approaches require labeled datasets of safe and unsafe states, and obtaining such labels is often not trivial because of non-obvious controlled forward invariant sets that are disjoint from failure sets, e.g., in the cases of systems with input constraints~\cite{howtotrain}, systems with complex observation-based failure sets that are hard to specify mathematically~\cite{latent_safety_filters_andrea_arXiv_2025}, or unknown failure states that the expert generating the demonstrations is implicitly avoiding.  
% It might be easier to generate a set of safe trajectories. 

We consider the setting where a safety-agnostic reference controller is available along with the system dynamics and a set of expert demonstrations that follow the reference controller while maintaining safety. Inverse constraint reinforcement learning (ICRL)~\cite{scobeeicl,icrl} or inverse constraint learning (ICL)~\cite{ICL,chou2018learning,chou2020learning,leung2023learning} consider the similar setting of a known reward function along with a simulator and a set of expert demonstrations. Corresponding algorithms infer constraint functions that explain the deviation of the expert trajectories from reward-maximizing behavior. % \cite{lindner2024learning} recently proposed an algorithm for learning constraints without the requirement of known rewards and dynamics. In this paper, we stick to the case with them known for simplicity.  
% \cite{BRT} have shown recently that the constraint function retrieved by an ideal ICL algorithm represents the backward reachable set of the set corresponding to some failure set. 
We adopt an ICL approach to learn a constraint function that we use to label a set of sampled trajectories which we then use to train a neural CBF. We call it an ICL-CBF.
% \yuxuan{in practical applications}, \yuxuan{synthesizing CBFs can be challenging or even infeasible}. 
% \yuxuan{Leveraging advancements in deep learning}, 
% \yuxuan{neural CBFs can be learned from offline datasets} % \cite{barriernet,sablas,howtotrain,vluefunctioniscbf}
%, which \yuxuan{alleviates the need for an explicit definition of safe states}. 
% \yuxuan{Nevertheless, this approach assumes that the training data is annotated, which is problematic since the safety set cannot be clearly defined.}
 %\yuxuan{These datasets are typically manually labeled, creating a significant scalability bottleneck for offline learning methods}. \yuxuan{Furthermore, label accuracy can be compromised, particularly when the boundary between safe and unsafe states is ambiguous}. For instance, \yuxuan{in scenarios where a vehicle trajectory results in a collision with a tree, it is difficult to pinpoint when the state transitions to unsafe, as the vehicle's braking distance is unspecified.}

\begin{figure}
    \centering
    \includegraphics[width=1\linewidth]{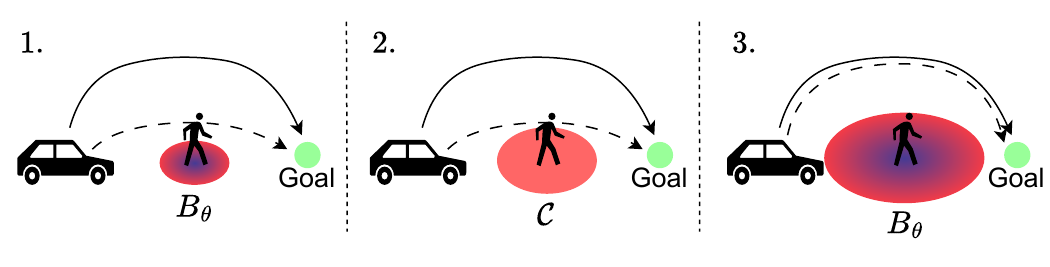}
    \caption{
    % \hussein{the text in the figure should be fixed: why the words inside the sentence capitalized? "Update C from Difference and Simulate" is not proper english. Maybe "Evaluate the difference between simulated and expert trajectory and update C"?} 
    The framework for training an ICL-CBF $B_\theta$, where the solid arrows represent the expert trajectories and the dashed ones represent the ones generated using a CBF-QP policy consisting of $\pi_{\mathit{ref}}$ and $B_\theta$. First, we train $B_\theta$ using trajectory data sampled with $\pi_\mathit{ref}$ that are labeled using the learned constraint function $\hat{c}_\phi$, and simulate the system while using it to filter unsafe actions. Then,  we compute the difference between the sampled trajectories and their corresponding expert ones and update $\hat{c}_\phi$ accordingly. Third, we retrain $B_\theta$ using a new set of sampled trajectories that are labeled using the new $\hat{c}_\phi$, simulate again, return to the second step, and repeat until convergence.}
    \label{fig:framework}
    \vspace{-0.5cm}
\end{figure}

We use a variation of MT-ICL~\cite{ICL}, an algorithm that uses  expert demonstrations from multiple tasks to learn a tight common constraint. Our proposed algorithm iteratively samples a set of trajectories following the reference controller, labels them using the constraint function being trained, trains a neural CBF using the labeled data, samples a new set of trajectories that follow the reference controllers while using a quadratic program based on the trained CBF as a safety filter, and retrains the constraint function to distinguish the sampled trajectories from the expert ones.  We visualize the procedure in Figure~\ref{fig:framework}.   We also propose a simple heuristic to accelerate training by postponing the training of the neural CBF until the last iteration. At that iteration, the constraint function would have converged and is able to provide more accurate labels. We evaluate the ability of learned ICL-CBFs to maintain  safety while minimally deviating from the reference controllers in four examples. We show that they outperform two baselines and achieve comparable results to those obtained when the ground-truth safety labels are available. 
%Our findings indicate that: (1) our method 
% is able to train a CBF that close to the upper bound CBF trained with annoated data
%\yuxuan{effectively trains a CBF closely matching the performance of an upper bound CBF trained with annotated data}, (2) 
% our method is able to recover unsafe states accurately
%\yuxuan{it accurately identifies unsafe states}, and (3) 
% our method achieve performance close to the upper bound with success/collision rates as the metrics
%\yuxuan{The proposed heuristic algorithm achieves better performance than the original method while costing less training time.} 

% The paper is organized as follows: Section \ref{sec:related_work} reviews related work. Sections \ref{sec:cbf} and \ref{sec:icl} cover the preliminaries of our method. We introduce our approach in Section \ref{sec:method}, followed by experimental results in Section \ref{sec:exp}. Finally, our conclusions are presented in Section \ref{sec:conclusion}.

\section{Related work}\label{sec:related_work}
\paragraph{Learning CBFs from expert demonstrations}
% \paragraph{Learning CBFs from offline data}
%Most existing methods for training neural CBFs   assume the availability of safe and unsafe trajectories  that are accurately  labeled~\cite{barriernet,sablas,clbf,vcbf}. 
%However, in many scenarios, such as when there are input constraints, high relative degree dynamics, or hard-to-specify failure sets, labeling trajectories as safe or unsafe is hard~\cite{howtotrain,latent_safety_filters_andrea_arXiv_2025}. 
% unsafe data is often hard to obtain or annotate
% acquiring and labeling unsafe data is challenging. 
%A state might not be in the failure set, but it might be in its  Backward Reachable Set (BRS), i.e., there is no policy that can guarantee the safety of the system starting from this state~\cite{time-dependent-Hamilton-Jacobi-formulation-Mitchell_2005}, and thus it should be considered unsafe. In other scenarios, generating trajectories that enter the failure sets, such as collisions in autonomous driving, is costly.  % Generating  expert demonstrations might be more feasible. 
Few algorithms have been proposed for learning neural CBFs from expert demonstrations~\cite{prevrocbf,rocbf,indcbf}. 
%\cite{chou2018learning,chou2020learning,leung2023learning,lindner2024learning}. %Recently, \cite{oodlabelcbf} 
% is proposed to alleviate labeling issue
%\yuxuan{attempt to address labeling challenges by using OOD analysis to label states}, 
% which uses Out-of-distribution (OOD) analysis to label states. 
%but it still needs enough annotated data to construct the OOD detector.
% \yuxuan{Despite this, they still require sufficient annotated data to construct an OOD detector.} 
%\yuxuan{Several methods} have been proposed to learn safety filters or constraints from expert trajectories %\cite{chou2018learning,chou2020learning,leung2023learning,lindner2024learning}. 
Robey et al. \cite{prevrocbf} introduced a method to learn neural CBFs solely from expert demonstrations by constructing a safe region encompassing their states, uniformly sampling  states around the region's boundary and labeling them as unsafe, and training the neural CBF with these sampled states along with the expert safe states. However, uniform sampling near the boundary region is often inefficient, especially in high-dimensional state spaces. In their follow-up work ROCBF \cite{rocbf}, the authors employ the reverse-KNN algorithm to detect the boundary of the safe region and consider  the states at that boundary as unsafe, thus avoiding sampling more states. 
% Another approach, iDBF \cite{indcbf}, also trains neural CBFs using only safe trajectories.
Casta\~neda et al. \cite{indcbf} proposed training neural CBFs using expert demonstrations  to avoid distribution shift calling them in-Distribution Barrier Functions (iDBFs). They train a Gaussian behavior cloning policy using the user-provided  demonstrations, then sample actions with low probability under that policy at each state in the demonstrations. At each such state and for each sampled action, they simulate the system and consider the resulting state as out-of-distribution, or equivalently from the view of the neural CBF training, unsafe. Instead, we exploit the knowledge of the reference controller to determine the states that were avoided by the expert due to safety violations, and consider them to be  unsafe. Farrel et al. \cite{oodlabelcbf}  recently proposed an approach to learn neural CBFs from an offline dataset that is a mix of labeled and unlabeled data, without further sampling. It utilizes out-of-distribution detectors for classifying the unlabeled data based on the labeled ones. 
% However, both ROCBF and iDBF \yuxuan{depend on the appropriate distribution of expert demonstrations} \yuxuan{and experience} performance degradation in \yuxuan{certain conditions}, which we will show in the experiments.

\paragraph{Inverse constraint learning}
Inverse Constraint Learning (ICL) was initially proposed in \cite{scobeeicl} as an analogue to Inverse Reinforcement Learning (IRL) \cite{maxent} for constraint inference, but it was limited to discrete settings. Malik et al. \cite{icrl} proposed ICRL that  generalizes ICL to continuous state spaces and model-free settings by learning  constraint functions. Kim et al. \cite{ICL} proposed MT-ICL and demonstrated that ICL can recover tighter constraints in the multi-task setting when the dynamics and the constraints are shared among the tasks. Building on MT-ICL, Qadri et al. \cite{BRT} proved that when the expert follows a maximum entropy policy or generates trajectories accomplishing multiple tasks sharing the same constraint, the constraint function inferred by an ideal ICL (or MT-ICL) algorithm
% A state might not be in the failure set, but it might be in its  Backward Reachable Set (BRS), i.e., there is no policy that can guarantee the safety of the system starting from this state~\cite{time-dependent-Hamilton-Jacobi-formulation-Mitchell_2005}
defines the Backward Reachable Set (BRS) corresponding to some unknown failure set, i.e., the set of states that for which there is no control policy that can prevent the system from eventually reaching the failure set in the worst case~\cite{time-dependent-Hamilton-Jacobi-formulation-Mitchell_2005}. 
% violate the dynamics-independent constraint. 
The complement of such BRS is the maximum controlled forward invariant set~\cite{worst_case_analysis_of_nonlinear_systems_Fialho_TAC_1999}.
% ~\cite{time-dependent-Hamilton-Jacobi-formulation-Mitchell_2005}. 
Thus, the constraint function obtained by an ideal ICL is a perfect classifier of safe and unsafe states. Based on a learned version of this classifier, we label a new set of sampled trajectories to train our neural CBFs.  
% , which means it depends on the system dynamics, \yuxuan{representing an optimal} form of CBF. 
%However, 
% normally, this assumption \hussein{it's not an assumption it is the goal of the algorithm, but it's not scalable} is hard to satisfy
% \yuxuan{satisfying this criterion is challenging and not typically scalable}. 
% \yuxuan{Unlike the aforementioned} methods, \yuxuan{our approach substitutes the reward function with a reference controller tailored for control tasks, and it imposes no specific requirements on expert demonstrations.}

\section{Preliminaries}
%\noindent \textbf{Notation}
%We denote by $\mathbb{R}$ the set of real numberes. 

We consider nonlinear control-affine systems of the form:
\begin{equation}\label{eq:system}
    \dot{x}=f(x)+g(x)u,
\end{equation}
where $x(t)\in \mathcal{X}\subseteq \mathbb{R}^n$ is the system's state and $u(t)\in \mathcal{U}\subseteq \mathbb{R}^m$ is the control input at time $t$. We assume that  $f:\mathcal{X}\rightarrow \mathbb{R}^n$ and $g:\mathcal{X}\rightarrow \mathbb{R}^{n\times m}$ are locally Lipschitz continuous and that system (\ref{eq:system}) is forward complete. 

\subsection{Control Barrier Functions}\label{sec:cbf}
% Control barrier functions usually serve as safety certificates in continuous-time nonlinear \hussein{control-affine} systems. 
  A CBF for system~(\ref{eq:system})  is  defined as follows.

\begin{definition}[Zeroing control barrier functions \cite{cbf}]
A continuously differentiable function $B: \mathcal{D}\rightarrow \mathbb{R}$ is a {\em zeroing control barrier function} for system~(\ref{eq:system}) if there exists an extended class-$\mathcal{K}_\infty$ function $\alpha:\mathbb{R}\rightarrow \mathbb{R}$, i.e., continuous, strictly increasing, and $\alpha(0)=0$, such that 
%it satisfies the following conditions:
\begin{equation}\label{eq:cbf}
    \begin{aligned}
      %  B(x)&\geq 0 \quad \forall x\in \mathcal{X}_{\mathit{safe}},\\
      %  B(x)&< 0 \quad \forall x\in \mathcal{X}\setminus \mathcal{X}_{\mathit{safe}}, \\
        \exists u\in \mathcal{U} \text{ s.t. } \dot{B}(x,u)+\alpha(B(x))&\geq 0 \quad \forall x\in \mathcal{D},\\
    \end{aligned}
\end{equation}
where $\dot{B}(x,u)=\nabla B(x) (f(x) + g(x) u)$ is the time derivative of $B$ along the trajectories of (\ref{eq:system}), and $\mathcal{D} = B_{\geq c} := \{x\ |\ B(x) \geq -c\}$, for some $c > 0$.
% is a superset for the super-level set $B_{\geq 0}:=\{x\ |\ B(x) \geq 0\}$.
%, and $\lim_{x\rightarrow \infty} \alpha(x)=\infty$.
\end{definition}

The super-level set $B_{\geq 0}$ can be made {\em forward invariant} by following a policy that satisfies (\ref{eq:cbf}). 
% is guaranteed to be forward invariant by following the subsequent policy:}

\begin{theorem}[\cite{cbf}]
Fix a CBF $B$ for system~(\ref{eq:system}) where $\nabla B(x) \neq 0$ for all $x\in B_{= 0} := \{x\ |\ B(x) = 0\}$. Then, any Lipshitz continuous control policy $\pi:\mathcal{X}\rightarrow \mathcal{U}$ that satisfies \begin{equation}\label{eq:cbf_controller}
    \pi(x)\in\{u\in\mathcal{U}:\nabla B(x)(f(x)+g(x)u)+\alpha(B(x))\geq 0\}
\end{equation}
will result in $B_{\geq 0}$ being 
% $\mathcal{X}_{\mathit{safe}}$ 
a {\em forward invariant set}, i.e., any trajectory of (\ref{eq:system}) starting in $B_{\geq 0}$ remains inside it $\forall t \geq 0$. 
\end{theorem}

% That is, given a controller $\pi$ that satisfies (\ref{eq:cbf_controller}), for any safe initial state $x_0\in \mathcal{X}_{\mathit{safe}}$, the state of the system will remain within $\mathcal{X}_{\mathit{safe}}$ under the control of $\pi$.

Then, given a reference controller $\pi_{\mathit{ref}}:\mathcal{D}\rightarrow \mathcal{U}$ and a failure set of states that is disjoint from $B_{\geq 0}$, a CBF $B$ for system~(\ref{eq:system}) can be used to design a {\em safety filter} that corrects $\pi_{\mathit{ref}}$'s decisions when violating the invariance of $B_{\geq 0}$, and thus maintaining safety. The safety filter can be constructed by formulating a quadratic program (QP) with the objective of finding a control input that is minimally distant from the reference one while belonging to the set defined in (\ref{eq:cbf_controller}) that guarantees the forward invariance of $B_{\geq 0}$ as follows~\cite{cbf}: 
% in the quadratic program for control:
\begin{equation}\label{eq:qp}
    \begin{aligned}
    \pi_{\mathit{safe}}(x) := \arg\min_{u\in \mathcal{U}}\lVert u-\pi_{\mathit{ref}}(x) \rVert, \\
        \ s.t.\ \nabla B(x)(f(x)+g(x)u)+\alpha(B(x))\geq 0. 
    \end{aligned}
\end{equation}

We call $\pi_{\mathit{safe}}$ a CBF-QP policy. 

\subsection{Inverse constraint learning}\label{sec:icl}
Compared to Inverse Reinforcement Learning (IRL), which aims to recover the reward function optimized for by an expert policy $\pi_{E}:\mathcal{D}\rightarrow\mathcal{U}$ in unconstrained settings,
% \pi_E\in \Pi$ to generate a set of trajectories, 
Inverse Constraint Learning (ICL) aims to recover the constraint that an expert policy in a constrained setting is satisfying while maximizing a known reward function given a set of trajectories generated by the expert policy.
% ICL has been studied in a series of previous works \cite{ICL,icrl, maxent}. 
%\yuxuan{
In this paper, we adopt the approach of Multi-Task ICL 
% the formulation presented in 
(MT-ICL) \cite{ICL}. In MT-ICL, the constraint of a single task is derived as the solution of the following constrained optimization problem: 
% \hussein{shouldn't this be flipped? the outer is learning the constraint and the inner is learning the policy?}
%}
\begin{align}
    \min_{\pi\in\Pi}J(\pi_E,r)-J(\pi,r)\  
    \text{s.t.}\  \max_{c\in \mathcal{C}}J(\pi,c)-J(\pi_E,c)\leq 0,
\end{align}
where $J(\pi,f)$ represents the value of policy $\pi$ under the measurement function $f$, $r$ is the known reward function of states, $\mathcal{C}$ is a set of constraint functions of states, and $\Pi$ is a set of policies. 
% In contrast to imitation learning that directly matches trajectories
%\yuxuan{Unlike imitation learning, which 
% directly 
%\hussein{aims to learn policies that mimic the actions taken in the expert trajectories}},
%matches trajectories}, 
%ICL the learned policy \yuxuan{is at least as safe as the expert's}. 
The optimization process comprises an outer part, which optimizes the constraint $\hat{c}$ as a classifier to maximally distinguish between the states visited by the learned policy $\hat{\pi}_E$  and those visited by the expert policy $\pi_E$:
\begin{equation}\label{eq:outer}
    \hat{c} = \mathop{\arg\max}_{c\in\mathcal{C}}J(\hat{\pi}_E,c)-J(\pi_E,c).
\end{equation}
The inner part trains a policy $\hat{\pi}_E$ 
% $\pi_i$ 
to maximize the reward given the current learned constraint $\hat{c}$:
\begin{equation}\label{eq:inner}
  \hat{\pi}_E = \max_{\pi\in\Pi}J(\pi,r),\quad\text{s.t. } J(\pi,\hat{c})\leq \kappa,
\end{equation}
for some user-defined cost threshold $\kappa$. 
The process can be seen as training a constraint that results in the same trajectories as the ones generated by a safe RL policy that maximizes the known reward while satisfying that constraint. 

Similarly, the constraint of $K$ tasks can be derived as the solution of the following optimization problem:
\begin{equation}
    \begin{aligned}
        \min_{\pi^{1:K}\in\Pi}\sum_i^K J(\pi^i_E,r^i)-J(\pi^i,r^i),\\
        \text{s.t.}\  \max_{c\in \mathcal{C}}\sum_i^KJ(\pi^i,c)-J(\pi_E^i,c)\leq 0,
    \end{aligned}
\end{equation}
where $\pi_E^i$, $\pi^i$, and $r^i$ are the expert policy, policy, and the reward function of task $i$. This version can be solved using the same strategy discussed above.

\section{Training neural CBFs using  ICL}\label{sec:method}

Qadri et al. \cite{BRT} proved that the constraint set, i.e., the set of states that are constraint-violating, derived through a single iteration of an {\em exact entropy-regularized} or multi-task ICL algorithm is the {\em backward reachable set (BRS)} corresponding to an unknown failure set. The complement of the % largest 
% infinite-time
BRS is the maximal controlled forward invariant set~\cite{robust_control_barrier_value_functions_choi_CDC_2021,worst_case_analysis_of_nonlinear_systems_Fialho_TAC_1999}.

In this paper, we adapt the ICL procedure to train a neural CBF from % unlabeled sampled trajectories and safe 
expert trajectories. 
% We assume that the expert policy is  a CBF-QP policy using a known $\pi_{\mathit{ref}}$ as a reference controller
% \yuxuan{
We approximate the expert policy by a CBF-QP policy with a known reference controller and a learned neural CBF. The expert does not need to be a CBF-QP policy. It can be a human or another safe task-achieving policy as long as the provided demonstrations sufficiently cover the state space besides the true constraint set. If the expert deviates from the provided reference controller within the safe set (the complement of the true constraint set), states visited by the CBF-QP policy with an ideal CBF $B^*$ whose sublevel set $B_{\geq 0}^*$ is the true constraint set, but not by the expert demonstrations, would be learned by ICL as part of the constraint set. Thus, the learned constraint set would be a conservative estimate of the true one.  

\subsection{Proposed algorithm}%Training the CBF with learned constraint}
% Since the learned constraint function $\hat{c}(x)$ \yuxuan{does not necessarily meet the conditions} in (\ref{eq:cbf}), it cannot be directly employed as a CBF. 
Given a learned neural CBF $B_\theta$ with parameters $\theta$ and the ground-truth reference controller $\pi_\mathit{ref}$, we define a corresponding CBF-QP policy $\hat{\pi}_{E}$, per (\ref{eq:qp}), as an approximation of the expert policy $\pi_{E}$. In contrast with (\ref{eq:inner}), we do not need to use a safe RL algorithm to obtain $\hat{\pi}_{E}$. Instead, we run a QP solver online to get the policy. On the other hand, as in MT-ICL, we sample trajectories using $\hat{\pi}_{E}$ and solve a similar optimization problem to  (\ref{eq:outer}) to find a constraint $\hat{c}_\phi$ that can be used to train $B_\theta$. 

Given a set of states visited by the expert trajectories $\mathcal{X}_E$ and a set of set of states visited by the sampled trajectories $\mathcal{X}_S^{c}$ using $\hat{\pi}_E$, we train a neural constraint function $\hat{c}_\phi$ parameterized with $\phi$ by  minimizing the following loss:
\begin{equation}\label{eq:c_update}
    \mathcal{L}_{\hat{c}_\phi}:=\sum_{x\in \mathcal{X}_S^{c}}\lVert 1 - \hat{c}_\phi(x)\rVert^2 + \sum_{x\in \mathcal{X}_E} \lVert 1 + \hat{c}_\phi(x)\rVert^2.
\end{equation}
% where we abuse notation and use $x \in \tau$ to refer to $x$ being a state visited in $\tau$. 
% We assume that $\tau$ here is a discrete-time trajectory that is a sampled-time version of a corresponding trajectory of system (\ref{eq:system}). 
% The recorded set of such states will be finite set in practice.
% instead of a continuous function of time.  
% We use this constraint to update $B_\theta$ and repeat the process.
% Nonetheless, it can be used to identify safe and unsafe states. 
% Given a set of sampled states $\mathcal{X}_S$, 
After that, we sample a new set of trajectories starting from random initial states using $\pi_{\mathit{ref}}$. We then partition the set of states in these trajectories, denoted by $\mathcal{X}_S^{B}$, to safe and unsafe ones:
% as follows: 
% we can define the sets of safe and unsafe states as:
% \begin{align}
$\mathcal{X}_{\mathit{safe}} = \{x\in \mathcal{X}_S^{B}\ |\  \hat{c}(x)<\delta\}$ and  
    % \label{eq:label_safe}\\
$\mathcal{X}_{\mathit{unsafe}}  = \{x\in \mathcal{X}_S^{B}\ |\ \hat{c}(x)\geq\delta\}$, 
    %,\label{eq:label_unsafe}
% \end{align}
where $\delta \in \mathbb{R}$ is a  hyperparameter. 
% user-specified classification threshold. 
Additionally, we define the set of safe state-action pairs, i.e., the ones where the state is in $\mathcal{X}_{\mathit{safe}}$ and the pair also  result in a state in $\mathcal{X}_{\mathit{safe}}$, in the sampled trajectories by 
% transitions
$\mathcal{D}_{\mathit{safe}}$. Once we obtain $\mathcal{X}_{\mathit{safe}}$, $\mathcal{X}_{\mathit{unsafe}}$,  and $\mathcal{D}_\mathit{safe}$, we train a neural CBF $B_\theta$ by minimizing the following loss:
\begin{equation}\label{eq:cbf_loss_function}
   \begin{aligned}
    \mathcal{L}_{B_\theta
    } :=&\ w_{\mathit{safe}}\sum_{x_{\mathit{safe}}\in \mathcal{X}_{\mathit{safe}}}\sigma(\epsilon_{\mathit{safe}}-B_{\theta}(x_{\mathit{safe}}))+\\  &w_{\mathit{unsafe}}\sum_{x_{\mathit{unsafe}}\in\mathcal{X}_{\mathit{unsafe}}}\sigma(\epsilon_{\mathit{unsafe}}+B_{\theta}(x_{\mathit{unsafe}}))\\
        +&w_{ascent}\sum_{(x_{\mathit{safe}},u_\mathit{safe})\in \mathcal{D}_{\mathit{safe}}}\sigma(\epsilon_{ascent}-\nabla B_{\theta}(x_{\mathit{safe}})\\
        &(f(x_{\mathit{safe}})+\ g(x_{\mathit{safe}})u_{\mathit{safe}})-\alpha \cdot B_\theta(x_{\mathit{safe}})),
    \end{aligned}
\end{equation}
where   $\sigma(\cdot)$ is the ReLU function and $\alpha$,  
%is an extended class-$\mathcal{K}_\infty$ function, and  
$w_{\mathit{safe}}$, $w_{\mathit{unsafe}}$, $w_{ascent}$, $\epsilon_{\mathit{safe}}$, $\epsilon_{\mathit{unsafe}}$, and $\epsilon_{ascent}$ are non-negative hyperparameters.
% are hyperparameters that weigh the importance of the different terms,  that control the desired CBF outputs under safe and unsafe states and controls. 
The training procedure is summarized in Algorithm \ref{alg:icl}.

\begin{algorithm}
    \caption{Training ICL-CBFs}\label{alg:icl}
    \KwIn{% Transition function $A(x,u)$, 
    reference controller $\pi_{\mathit{ref}}$, a set of expert demonstrations $\{\tau_E\}$, and system dynamics.}
    Initialize $\hat{c}_\phi$ and $B_\theta$ as zero functions.\\
        \For{$i$ in $1,\dots,N$}{
        Generate $\mathcal{X}_S^{B}$ by sampling trajectories using  $\pi_{\mathit{ref}}$  starting from random initial states.\label{ln:sample_using_pi_ref}\\
        Construct $\mathcal{X}_\mathit{safe}$, $\mathcal{X}_\mathit{unsafe}$, and $\mathcal{D}_\mathit{safe}$ by labeling the sampled trajectories using $\hat{c}_\phi$. \label{ln:x_safe}\\ %  with (\ref{eq:label_safe}-\ref{eq:label_unsafe})\\
        Train $B_\theta$ according to (\ref{eq:cbf_loss_function}).\label{ln:train_b}\\
        \If{$i=N$} {\bf break}
        Generate $\mathcal{X}_S^{c}$ by sampling trajectories using $\hat{\pi}_E$ \label{ln:sample_using_pi_E_hat}
        %the CBF-QP policy with $\pi_\mathit{ref}$ and $B_\theta$ 
        starting from random initial states. \\
        Update $\hat{c}_\phi(x)$ using and according to (\ref{eq:c_update}).\label{ln:update_c}
    }
    \KwOut{neural control barrier function $B_{\theta}$.}
\end{algorithm}

\begin{remark}
    Algorithm~\ref{alg:icl} can be extended straightforwardly to accept multi-task expert demonstrations and corresponding reference controllers as input as follows: in line \ref{ln:sample_using_pi_ref}, it generates a union $\mathcal{X}_S^B$ of sets $\{\mathcal{X}_S^{B,i}\}$ by sampling trajectories using $\pi_{\mathit{ref}}^i$ for each task $i$; then in line~\ref{ln:sample_using_pi_E_hat}, it generates a union $\mathcal{X}_S^c$ of sets $\{\mathcal{X}_S^{c,i}\}$, one for each task $i$ using its corresponding CBF-QP policy $\hat{\pi}_E^i$ resulting from $\pi_\mathit{ref}^i$ and $B_\theta$; we train $B_\theta$ with $\{\mathcal{X}_S^{B}\}$ in line \ref{ln:train_b} and update $\hat{c}_\phi$ using $\{\mathcal{X}_S^{c}\}$ in line \ref{ln:update_c}.
\end{remark}

\subsection{Heuristic for approximating $\hat{\pi}_E$ during training}% Recovering the constraint with ICL} 

Retraining the neural CBF $B_\theta$ every iteration of Algorithm~\ref{alg:icl} can be computationally expensive. When the action space is low-dimensional, it might be more efficient to use a grid search to find a safe control during training. For that reason, in our experiments, in the first $N-1$ iterations, we grid the control space over which we search for a control input that does not violate the constraint $\hat{c}_\phi$ to generate $\hat{\pi}_E$, 
% that follows $\pi_{\mathit{ref}}$ while satisfying $\hat{c}_\phi$,
and only train $B_\theta$ in the last iteration. 
% we only train the constraint function $\hat{c}_\phi$. 

To generate $\hat{\pi}_E$, we select a ball centered at the origin in  the control space $\mathcal{U}$ and grid it according to a user-defined resolution. We also select a sampling time $t_s$. Then, we define $\hat{\pi}_E$ to be the policy that  samples the state $x$ every $t_s$ seconds. At each sampling instant, $\hat{\pi}_E$  iterates over the cells of the grid.  For each cell, it computes the distance of its center to $\pi_{\mathit{ref}}(x)$ at the current state. Then, it simulates the system by fixing the control signal to be a constant one that is equal to the center of the cell for $t_s$ seconds.  Finally, it selects the closest center to $\pi_{\mathit{ref}}(x)$ that does not lead to a state $x'$ that violates the learned constraint $\hat{c}_\phi$ after $t_s$ seconds, i.e.,  the case where $\hat{c}_\phi(x') \geq \delta$.

This heuristic allows skipping sampling a new set of trajectories, labeling them using $\hat{c}_\phi$, and training the CBF $B_\theta$ at each iteration in Algorithm~\ref{alg:icl}. Once the constraint converges, we can use it to train a neural CBF that can be used to filter unsafe actions during deployment. 
The neural CBF would then allow the design of the CBF-QP policy that is more computationally efficient to run in real-time. 
%\yuxuan{
 Note that this heuristic is only useful for 
% applies only to 
systems with low-dimensional action spaces. As the action space dimension increases, 
% When they have high-dimensional ones instead, 
the original algorithm becomes more efficient. 
% remains appropriate, 
%because in these settings, trajectory sampling, rather than neural-network training, constitutes the primary computational bottleneck. }
%That is because the one used during training of $\hat{c}_\theta$ requires gridding the action space and sequentially iterating over the grid cells.
% However, the latter would still be more efficient than retraining the CBF every iteration of Algorithm~\ref{alg:icl}. 

\section{Experimental results}\label{sec:exp}
We evaluated our method in four scenarios aiming 
% .}
% now empirically demonstrate that when replacing CRL with our approach, the trained $\hat{c}_\phi$ can correctly label states to train a CBF that guarantees safety.  
% \yuxuan{ Our aim is 
to address the following research questions (RQs):
\begin{itemize}
\item \textbf{RQ1:} Do ICL-CBF-based safety filters improve safety while minimally affecting task success? 
% How closely do they align with the ground truth labels?
\item \textbf{RQ2:}  Does the heuristic algorithm for approximating $\hat{\pi}_E$ degrade the quality of $B_\theta$? 
\item \textbf{RQ3:} Should we generate $\mathcal{X}_S^B$ using $\pi_\mathit{ref}$ or $\hat{\pi}_E$?
\item \textbf{RQ4:} Are the safety labels produced using the learned constraint accurate?
\yuxuan{
\item \textbf{RQ5:} How sensitive is the performance of the learned ICL-CBF  to the choice of the hyperparameter $\delta$? %and can $\delta$ be selected with proper effort?
}
\end{itemize}

\subsection{Setup}
%To evaluate the performance of our method, we conduct a series of experiments in 
The scenarios we consider are:
%\begin{itemize}
%\item 
\textbf{Single integrator}: driving a simple 2D robot with single integrator dynamics to reach a  goal position while avoiding a circular obstacle; 
% \item 
\textbf{Dubins car}: same as the first scenario  but with Dubins car dynamics~\cite{Dubins_car}; 
    % Maneuvering the \yuxuan{Dubins} car toward the goal position while preventing collisions with a square obstacle located at the origin.
    % \item 
    \textbf{Inverted pendulum \cite{invertedpendulum}}: rotating an inverted pendulum according to a reference controller while avoiding an unsafe set of states;
    % it remains within the defined safe set.
    %\item 
    \textbf{Quadrotor \cite{quadrotor}}: navigating a quadrotor towards a goal position while avoiding colliding  with the ground.
% \end{itemize}

We compare our results with those obtained using the ground-truth CBFs (GT) (which we used to generate the expert demonstrations), the neural CBFs trained using ROCBF \cite{rocbf},  and the neural CBFs trained using iDBF \cite{indcbf}.
%\yuxuan{Since ROCBF is designed for systems with partially accurate dynamics, and our experiment assumes fully accurate dynamics,}
%% Since ROCBF also assumes the dynamics are partially accurate, we assume the dynamics are accurate. 
% we only implement the data generation part proposed by ROCBF to train CBFs. 
%% Besides, we train CBFs (L-CBF) according to the same loss in (\ref{eq:cbf_loss_function}) using states annotated with safety labels generated by GT as the upper bound.
% \yuxuan{
We also compare them with those obtained using the neural CBFs trained using a set of sampled trajectories using the reference controllers whose states are labeled as safe or unsafe based on  the ground-truth CBF, which we denote by L-CBF (L stands for ``labeled'').
% , and using the same loss function in (\ref{eq:cbf_loss_function}).
% , employing states labeled with safety annotations from GT.} 
For all scenarios, we use multi-layer perceptrons (MLPs) to represent the constraint functions and the neural CBFs. We choose the hyperparameter $\delta$  
% for the constraint function  
to be $0.6$ for the single integrator and the quadrotor, $0.4$ for the Dubins car, and $0.3$ for the inverted pendulum.

\subsection{Implementation Details}
\label{sec:implementation_details}
%\yuxuan{
% During implementation, 
% We use a CBF-QP policy that leverages ground-truth information as the expert policy, because it follows the reference controller as closely as possible while ensuring safety. 
We designed CBFs and reference controllers to generate the expert demonstrations.  However, generally, the expert policy need not be a CBF-QP policy, as we mentioned earlier.
%in Section~\ref{sec:method}. }
% Expert demonstrations can also be collected by a human or another safe task-achieving policy, provided it follows the reference controller when it is not safety-violating.
\subsubsection{Single integrator}
\label{app:single_integrator}
% We start with a simple Single integrator task, which aims to reach the goal position while avoiding a circle obstacle positioned at the origin.
This scenario consists of a 2D  robot with single integrator dynamics reaching a specified goal position while avoiding a circular obstacle located at the origin. Given the state $\boldsymbol{\mathrm{x}}=[x,y]^\top$, the robot's dynamics are defined as $\dot{\boldsymbol{\mathrm{x}}} = u = [v_x, v_y]^\top$.
The   failure set of states is defined as the circle centered at the origin with radius $r$, i.e., $\{\boldsymbol{\mathrm{x}}\in \mathbb{R}^2\ |\ \lVert \boldsymbol{\mathrm{x}}\rVert_2<r\}$, where $r=1$ is the radius of the obstacle. We sample initial states  uniformly at random from the square $[-6,6]\times[-6,6]$ while rejecting those with $\|\mathrm{x}\|_2 < 3$. For each initial state $\boldsymbol{\mathrm{x}}_0=[x_0,y_0]^\top$, we set the goal state to be $\boldsymbol{\mathrm{x}}_g=[-x_0+b_x,-y_0+b_y]^\top$, where $b_x,b_y$ are sampled uniformly at random  from the interval $[-1,1]$. 
% Such a choice drives to ensure the agent will always step into the unsafe states with 
We define the reference controller for the goal $x_g$ to be $\pi_{\mathit{ref}}(\boldsymbol{\mathrm{x}}, \boldsymbol{\mathrm{x}}_g):=\frac{\boldsymbol{\mathrm{x}}_g-\boldsymbol{\mathrm{x}}}{\lVert \boldsymbol{\mathrm{x}}_g-\boldsymbol{\mathrm{x}}\rVert_2}$. Such a choice of initial and goal states drives the agent to pass close to the origin since  they are at opposing ends. Thus, the agent will enter the failure set when following the reference controller if not prevented using a safety filter. 
To collect expert demonstrations, we used a CBF-QP controller with the same reference controller and the CBF
%\begin{equation}
$B_{GT}(\boldsymbol{\mathrm{x}}) = \lVert\boldsymbol{\mathrm{x}}\rVert_2 - r$. We considered the agent to have reached the goal when it is within 0.1 distance. The maximum number of time steps in each trajectory is 300 at which it is terminated if it did not reach the goal. We collected 150 trajectories with different initial and goal states for training. The sampling time is set to 0.1 seconds for training and evaluation. For the grid-based $\hat{\pi}_E$, we constructed a $50\times 50$ grid over the square $[-1,1] \times [-1,1]$ in the control space.
% we uniformly select 50 values ranging from $[-1,1]$ for each dimension of the action during prediction.

\subsubsection{Inverted pendulum}
We followed the same setting used in \cite{Verification_and_Synthesis_using_SoS_Andrew_Clark_CDC_2021}, where the state is $\boldsymbol{\mathrm{x}} = [x,y]$ and  
the dynamics are defined as  follows: $\dot{\boldsymbol{\mathrm{x}}}=\begin{pmatrix}
        0 & 1 \\
        1 & 0
    \end{pmatrix}\boldsymbol{\mathrm{x}} + \begin{pmatrix}
        0\\
        1
    \end{pmatrix}u$.
The failure set is the complement of the set  $[-0.1,0.15]\times[-0.3,0.25]$, the reference controller is $\pi_{\mathit{ref}}(\boldsymbol{\mathrm{x}})=-K\boldsymbol{\mathrm{x}}$ where $K=[3, 3]$, and the CBF used in the CBF-QP policy that we used to generate the expert demonstrations  is 
%\begin{equation}
    $B_{GT}(\boldsymbol{\mathrm{x}})=k-(\boldsymbol{\mathrm{x}}-\boldsymbol{\mathrm{x}}_g)^\top P(\boldsymbol{\mathrm{x}}-\boldsymbol{\mathrm{x}}_g)$,
%\end{equation}
where $P=\begin{pmatrix}
    1.25 & 0.25\\
    0.25 & 0.25
\end{pmatrix}$, $k=0.01$, and $\boldsymbol{\mathrm{x}}_g=[0,0]$ . We 
%use the CBF-QP policy consisting of  $\pi_{\mathit{ref}}(\boldsymbol{\mathrm{x}})$ and $B_{GT}$ to collect trajectories starting from states 
sampled initial states uniformly at random from the rectangle  $[-0.103,0.148]\times[-0.3,0.25]$. We considered the agent to have reached the goal (origin) when its norm is less than or equal to 0.1. The maximum number of time steps in each trajectory is 300 at which it is terminated if it did not reach the goal. The sampling time is 0.1 seconds. For the grid-based $\hat{\pi}_E$, we split the interval $[-5,5]$ into 2500 equal intervals.
%we uniformly select 2500 values ranging from $[-5,5]$ for each dimension of the action during prediction.

\subsubsection{Dubins car}
We followed a similar setting used in \cite{quadrotor}, where the state is $\boldsymbol{\mathrm{x}}=[x,y,\phi]^\top$ and the system dynamics are: $\dot{\boldsymbol{\mathrm{x}}} = \begin{pmatrix}
        v\cos(\phi)\\
        v\sin(\phi)\\
        u
    \end{pmatrix}$,
where $v=1$ is the fixed speed of the car. The task is to drive the car starting from initial states sampled uniformly at random from the square with sides equal to $0.4$ and centered at  $(-3,-3)$ for the 2D position  and the range $[-0.4\pi,0.4\pi]$ for the heading angle to the goal position $(x_g,y_g)=(3,3.5)$, while avoiding the failure set defined by the square centered at the origin with sides equals to 2. We used $\pi_\mathit{ref}(\mathrm{x}):=\psi-\phi$ as the reference controller, where $\psi=\arctan(y_g-y,x_g-x)$. We trained a neural CBF using the method proposed in \cite{quadrotor} to generate 100 expert trajectories 
% . We collected 100 trajectories
with different initial states.
% for training. 
We considered the car to have reached the goal when it is within 0.1 distance. The maximum number of time steps in each trajectory is 200 at which it is terminated if it did not reach the goal. The sampling time is 0.1 seconds. For the grid-based $\hat{\pi}_E$, we split the interval $[-1,1]$ into 50 equal intervals.
% we uniformly select 50 values ranging from $[-1,1]$ for each dimension of the action during prediction.

\subsubsection{Quadrotor}
We considered a self-righting 6-state planar quadrotor model used in \cite{quadrotor} with a state $\mathrm{x}=[x,v_x,y,v_y,\phi,\rho]^\top$ and dynamics: 
\begin{equation}
    \dot{\boldsymbol{\mathrm{x}}}=\begin{pmatrix}
        v_x\\
        \frac{-C_D^v v_x}{m}\\
        v_y\\
        \frac{-C_D^v v_y}{m}-g\\
        \rho\\
        \frac{-C_D^\phi \rho}{I_{yy}}
    \end{pmatrix} + \begin{pmatrix}
        0 & 0\\
        \frac{-\sin(\phi)}{m} & \frac{-\sin(\phi)}{m}\\
        0 & 0\\
        \frac{\cos(\phi)}{m} & \frac{\cos(\phi)}{m}\\
        0 & 0 \\
        -\frac{l}{I_{yy}} & -\frac{l}{I_{yy}}
    \end{pmatrix}u.
\end{equation}
The task is to drive the quadrotor with initial states of the form $\mathrm{x}=[0,1,2+b_y,0,-\frac{\pi}{2.5}+b_\phi,0]$, where $b_y$ and $b_\phi$ are sampled uniformly at random from the ranges $[-0.1,0.1]$ and $[-0.05,0.05]$, respectively, 
%is  sampled uniformly at random from the range , 
to the goal position (6,9), while avoiding the failure set which consists of all the states with $y = 0$, representing the ground. We used the LQR-based controller considered in \cite{quadrotor} as the reference controller $\pi_\mathit{ref}$. We trained a neural CBF using the method proposed in \cite{quadrotor} to generate the expert trajectories. We collected 60 trajectories with different initial states for training.  We considered   the quadrotor to have reached the goal when its within 0.1 distance. The maximum number of time steps in each trajectory is 300 at which it is terminated if it did not reach the goal. The sampling time is set to 0.05 seconds. For the grid-based $\hat{\pi}_E$, we constructed a $100\times 100$ grid over the square $[0,20] \times [0,20]$ in the control space.   %for each dimension of the action during prediction.
\subsection{RQ1: Do ICL-CBF-based safety filters improve safety while minimally affecting task success? }
\begin{figure}
    \centering
    \subcaptionbox{Learned constraint\label{subfig:learned_constraint}}{
    \includegraphics[width = .4\linewidth]{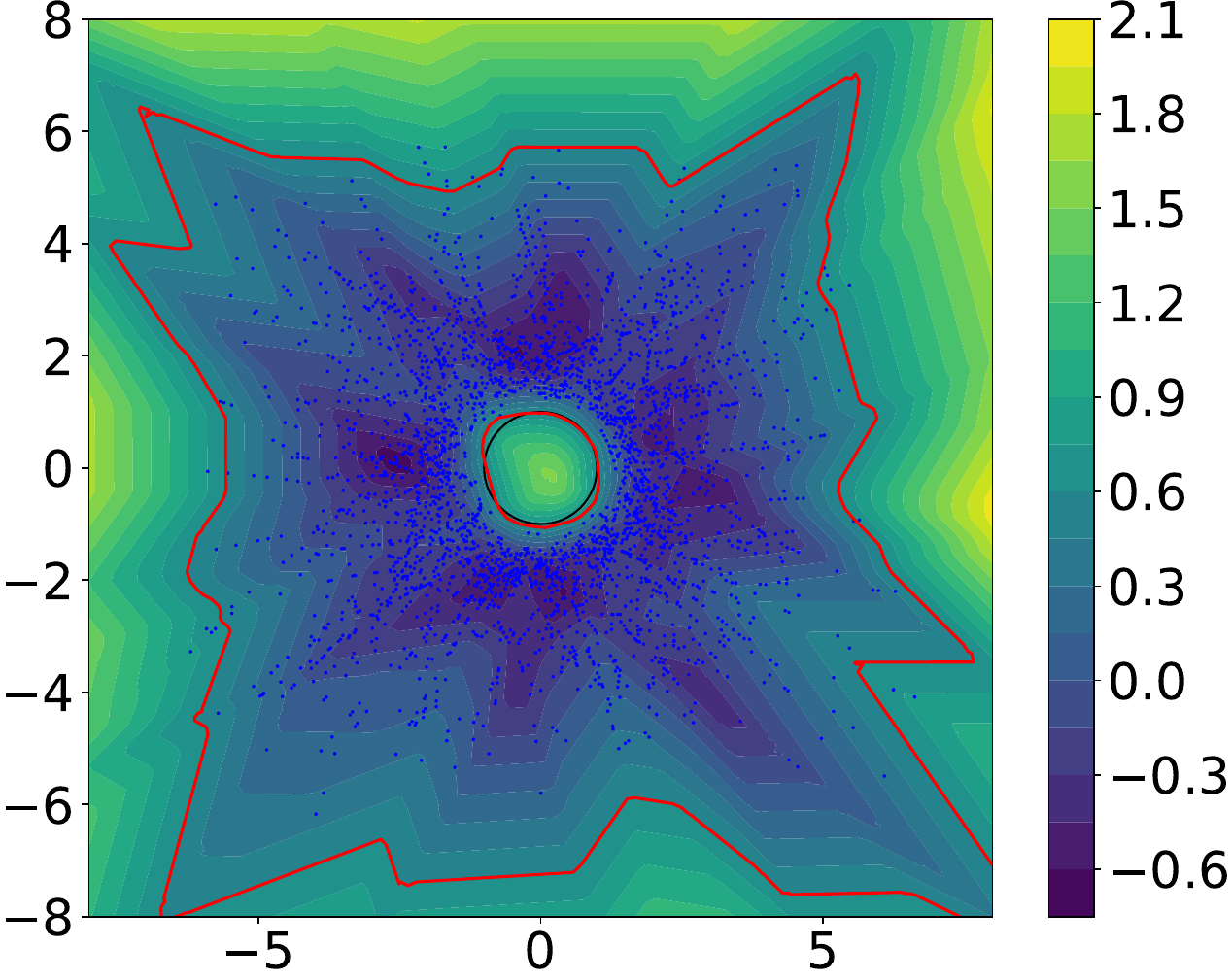}
    }
    \subcaptionbox{ICL-CBF\label{subfig:icl_cbf}}{
    \includegraphics[width = .4\linewidth]{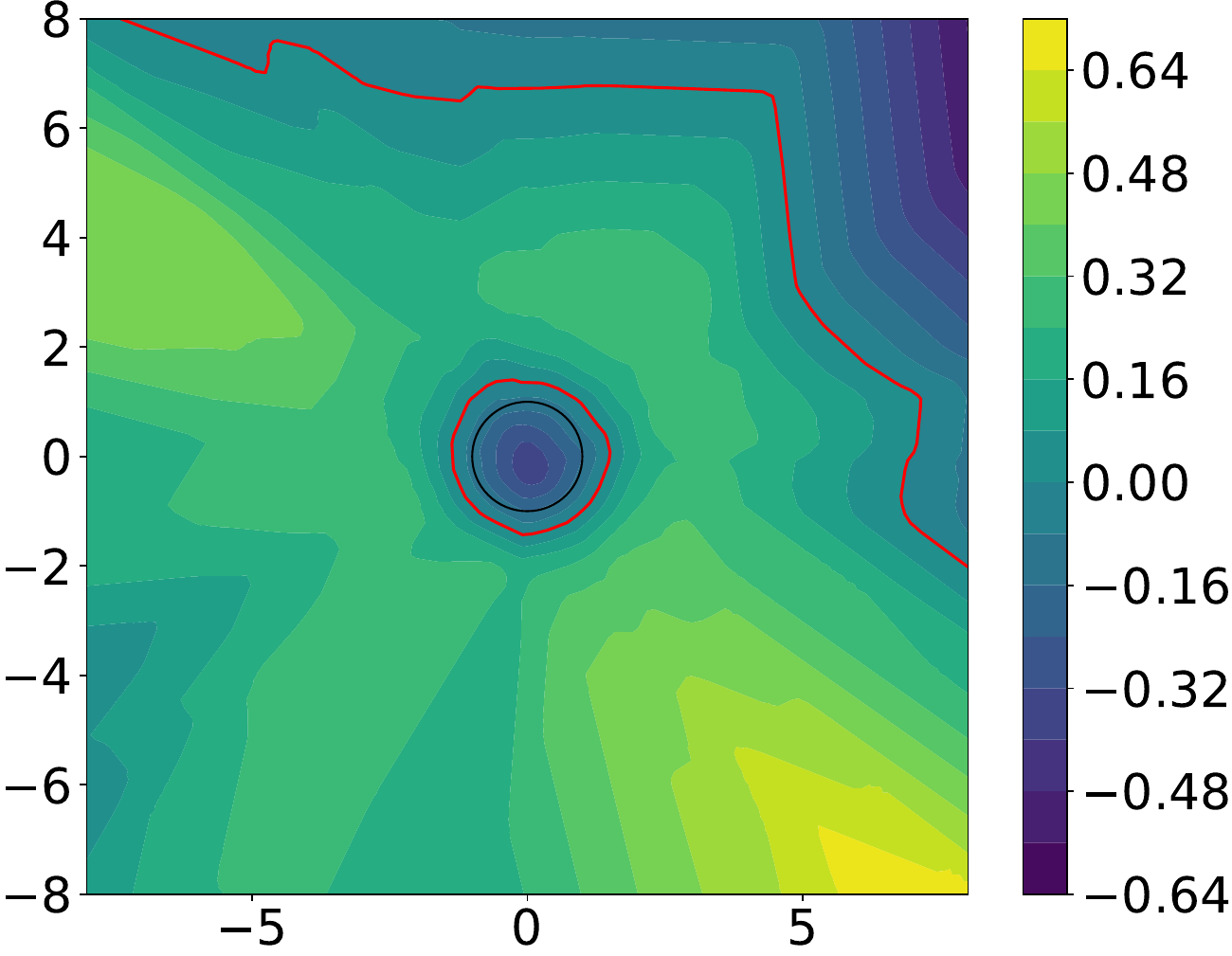}
    }
    \subcaptionbox{iDBF\label{subfig:idbf}}{
    \includegraphics[width = .4\linewidth]{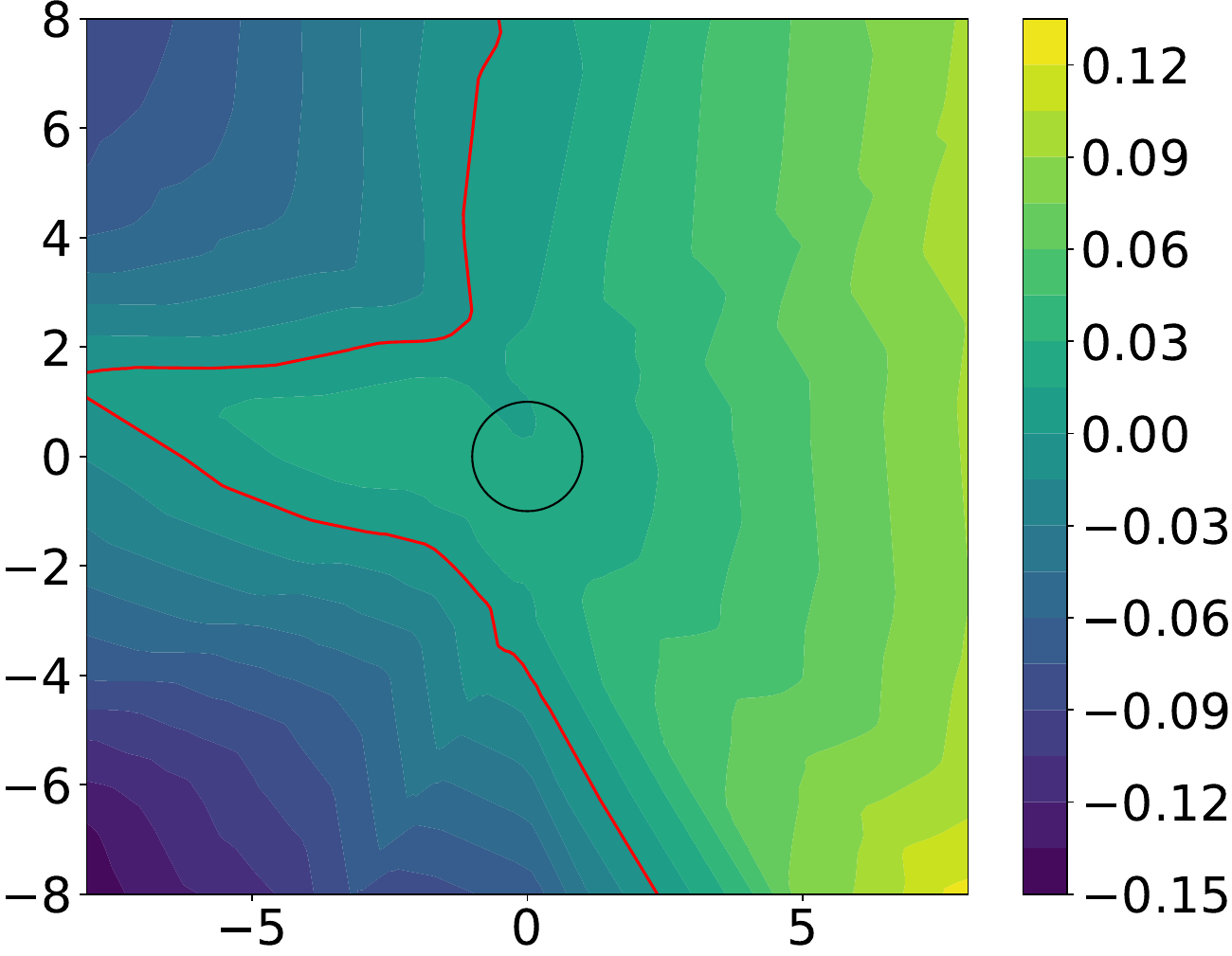}
    }
    \subcaptionbox{ROCBF\label{subfig:rocbf}}{
    \includegraphics[width = .4\linewidth]{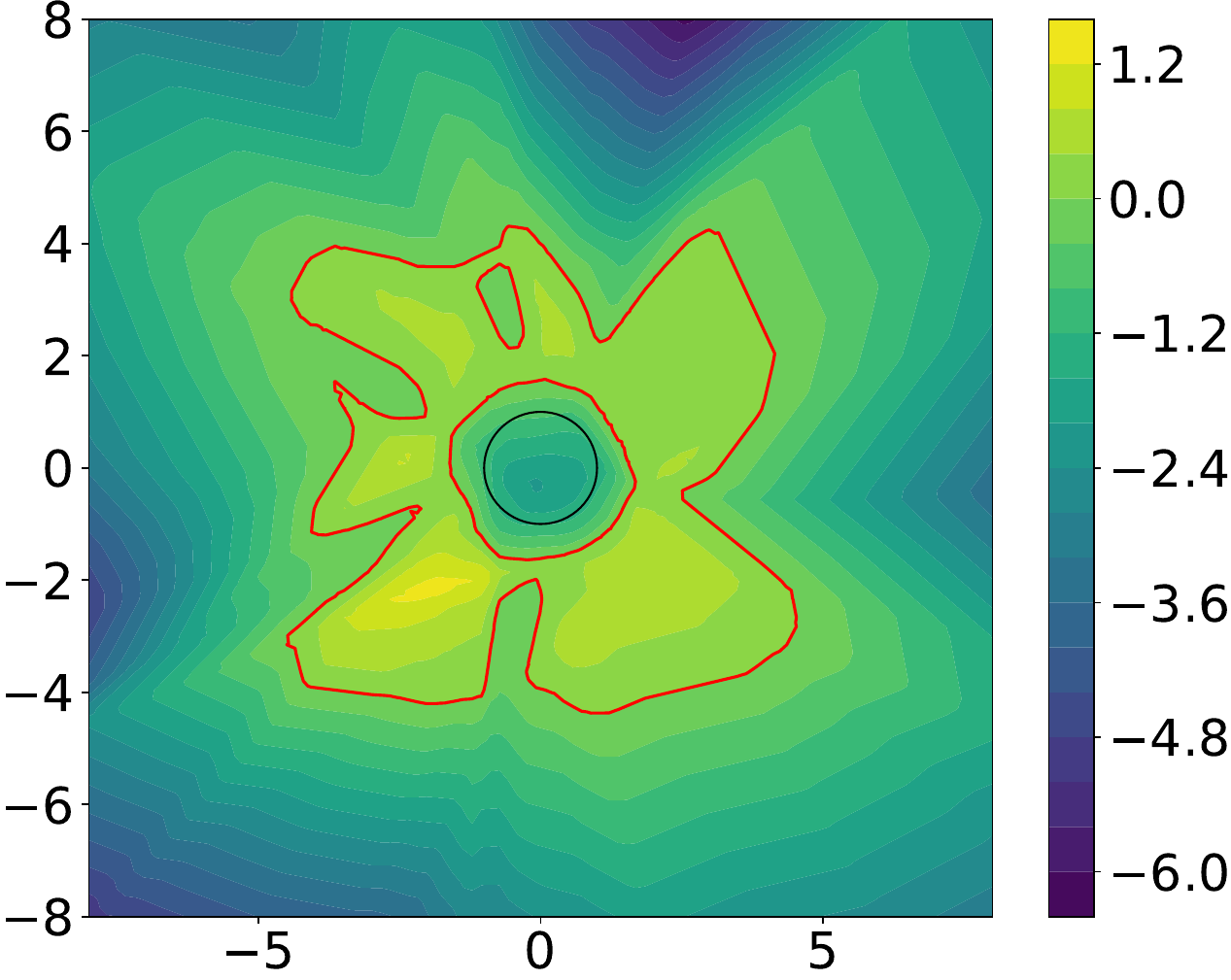}
    }
    \subcaptionbox{L-CBF\label{subfig:cbfl}}{
    \includegraphics[width = .4\linewidth]{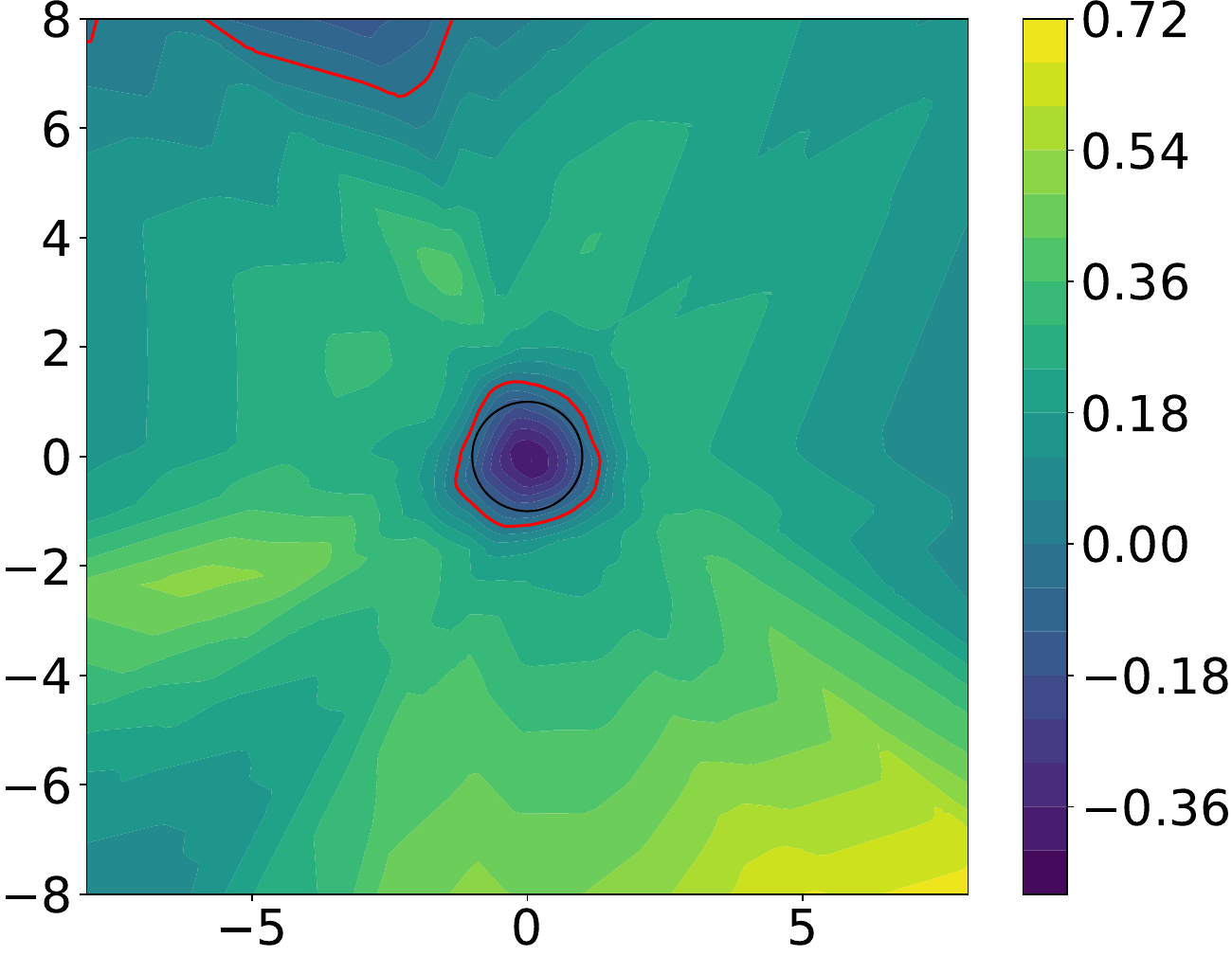}
    }
    \subcaptionbox{Ground truth\label{subfig:gt}}{
    \includegraphics[width = .4\linewidth]{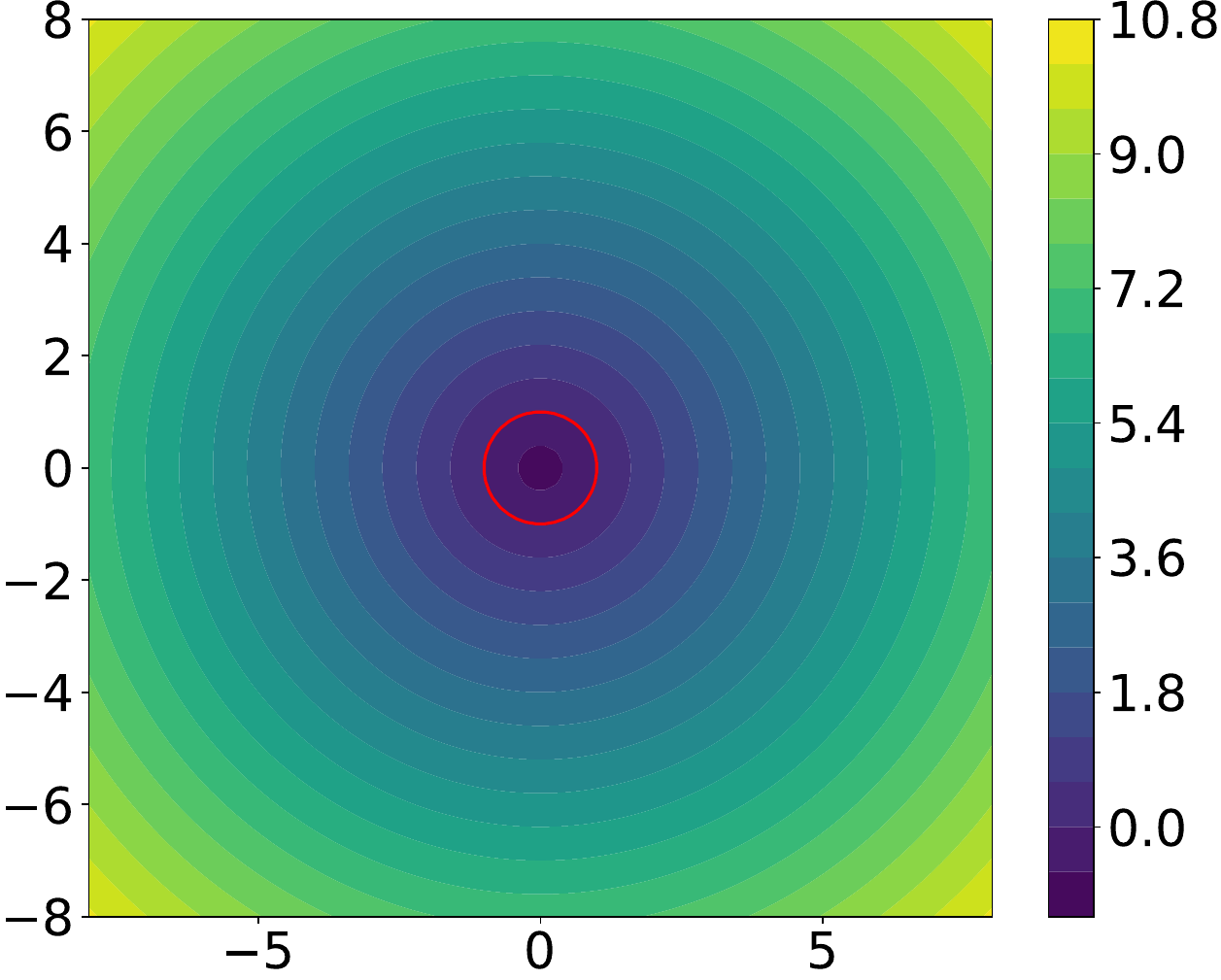}
    }
    % \hussein{the font of the numbers in the figures should be increased.}
    \caption{ Values of the learned constraint and neural CBFs in the single-integrator scenario, where the black circles in the middle of the figures represent the obstacle. For (a), we color the boundary of the set $\{\boldsymbol{\mathrm{x}}\in\mathbb{R}^2|\hat{c}_\phi(\boldsymbol{\mathrm{x}})=\delta\}$ in red and the blue points are randomly sampled states from the expert demonstrations. For (b)-(f), the red margins denote the zero level set of each (neural) CBF. \label{fig:single_integrator_cbfs}}
    \vspace{-0.5cm}
\end{figure}

We first subjectively analyze the quality of the  constraints and the neural CBFs  learned using our method for the single integrator case before analyzing their closed-loop performances. We visualize them in Figure~\ref{fig:single_integrator_cbfs}. 
%We first examine the accuracy of the recovered unsafe regions in the single integrator example.
In Figure \ref{subfig:learned_constraint}, 
% the area surrounding 
we can see that the obstacle is correctly identified as part of the constraint set by having corresponding learned constraint function's values greater than $\delta$. Note that because the system can stop at any state by choosing a zero control input, 
% of the absence of input constraints, 
the backward reachable set is equal to the failure set, i.e., the states in the obstacle. That makes a simple distance-based function a valid CBF, which is the one we used to generate the expert trajectories, as we explained in Section~\ref{sec:implementation_details}. 
% That means that with a sufficiently large and diverse set of expert trajectories, the learned constraint by ICL should be the obstacle itself~\cite{BRT}.   
In our case, since the expert trajectories are concentrated around the obstacle, the region of the state space further away is also considered constraint-violating by the learned constraint, as expected. 
It is evident from Figure \ref{subfig:idbf} that iDBF did not accurately recover the failure set. That is likely because of its strategy for sampling  states to label as out-of-distribution (or equivalently, unsafe), which in low-dimensional state spaces and in the presence of dense set of expert trajectories result in overlapping regions labeled both safe and unsafe, significantly affecting the quality of the learned constraint.
% , confusing the training procedure. 
Although ROCBF successfully delineates the failure set, part of the region visited by the expert trajectories is  misclassified as unsafe, as shown in Figure~\ref{subfig:rocbf}. In contrast, ICL-CBF and L-CBF, as shown in Figures~\ref{subfig:icl_cbf} and \ref{subfig:cbfl}, respectively, are very close to each other implying that training a neural CBF from the labels generated by a learned constraint using ICL is comparable to training it using the true labels. 
% Only the zero level set of $B_\theta$ effectively separates the safe from unsafe regions. 
% Additionally, $B_\theta$ closely resembles the CBF learned with annotated data, demonstrating that our method can achieve results comparable to the upper bound.
%, demonstrating that our method effectively identifies the unsafe states. 
% Moreover, the states not visited by the expert trajectories are also  assigned high constraint function values, and thus also identified as constraint-violating. 
% This is exbecause ICL add the
% These are states visited by the sampled trajectories during the ICL procedure.
% to the constraint set.
%is the out-of-distribution set of states of the expert set. 
% Ideally, when more expert demonstrations are introduced and eventually cover all the safe states, the learned constraint set will progressively recover the true constraint set. 
% The impact of these states can be disregarded since they are not encountered in either the expert trajectories or the trajectories generated by the reference controller. Therefore, we claim that the unsafe regions identified by our method are accurate.

%We then evaluate the quality of CBF trained using ICL-CBF. 

Second, we compare the closed-loop performance of the neural CBFs trained using the different methods based on the collision rates (CR) and success rates (SR) when used as part of the QP-based safety filters. The results are shown in Table \ref{tab:performance}. ICL-CBF achieves the best performance in terms of both metrics in all considered scenarios. For the single integrator one, although ROCBF results in a zero CR, it results in a low SR due to its conservativeness. Compared to L-CBF, ICL-CBF results in a $5.6\%$ decrease in SR in the single integrator scenario and a $20.8\%$ decrease in SR in the quadrotor one. For the inverted pendulum  and Dubins car scenarios, ICL-CBF achieves comparable performance to L-CBF. In contrast, iDBF results in high CR in both tasks. 
% , likely due to the dense distribution of expert demonstrations, which hinders its data sampling strategy, as mentioned in Appendix \ref{sec:data}.

\begin{figure}
    \centering
    \subcaptionbox{ICL-CBF\label{subfig:traj_icl}}{
    \includegraphics[width = .4\linewidth]{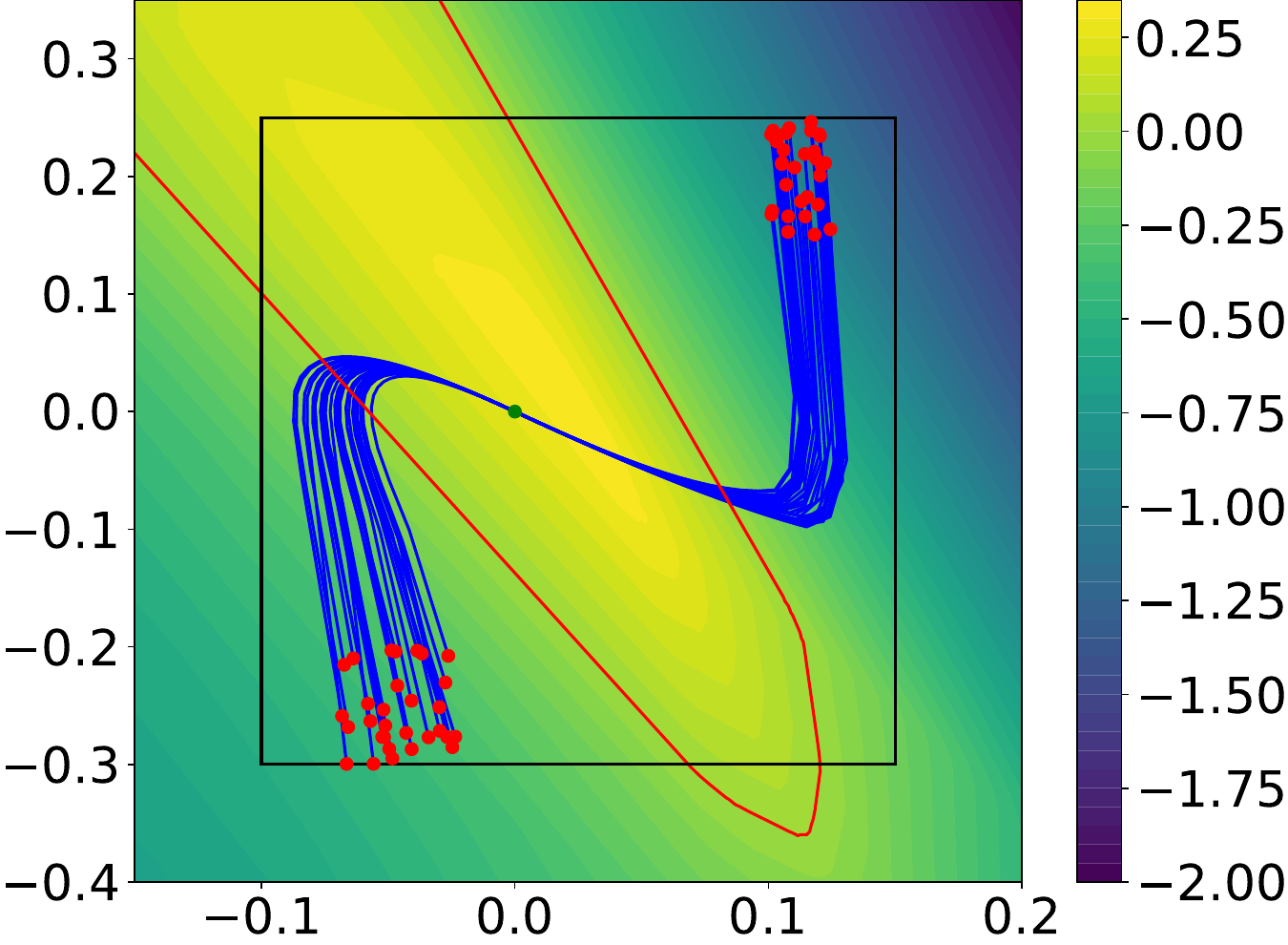}
    }
    \subcaptionbox{iDBF\label{subfig:traj_idbf}}{
    \includegraphics[width = .4\linewidth]{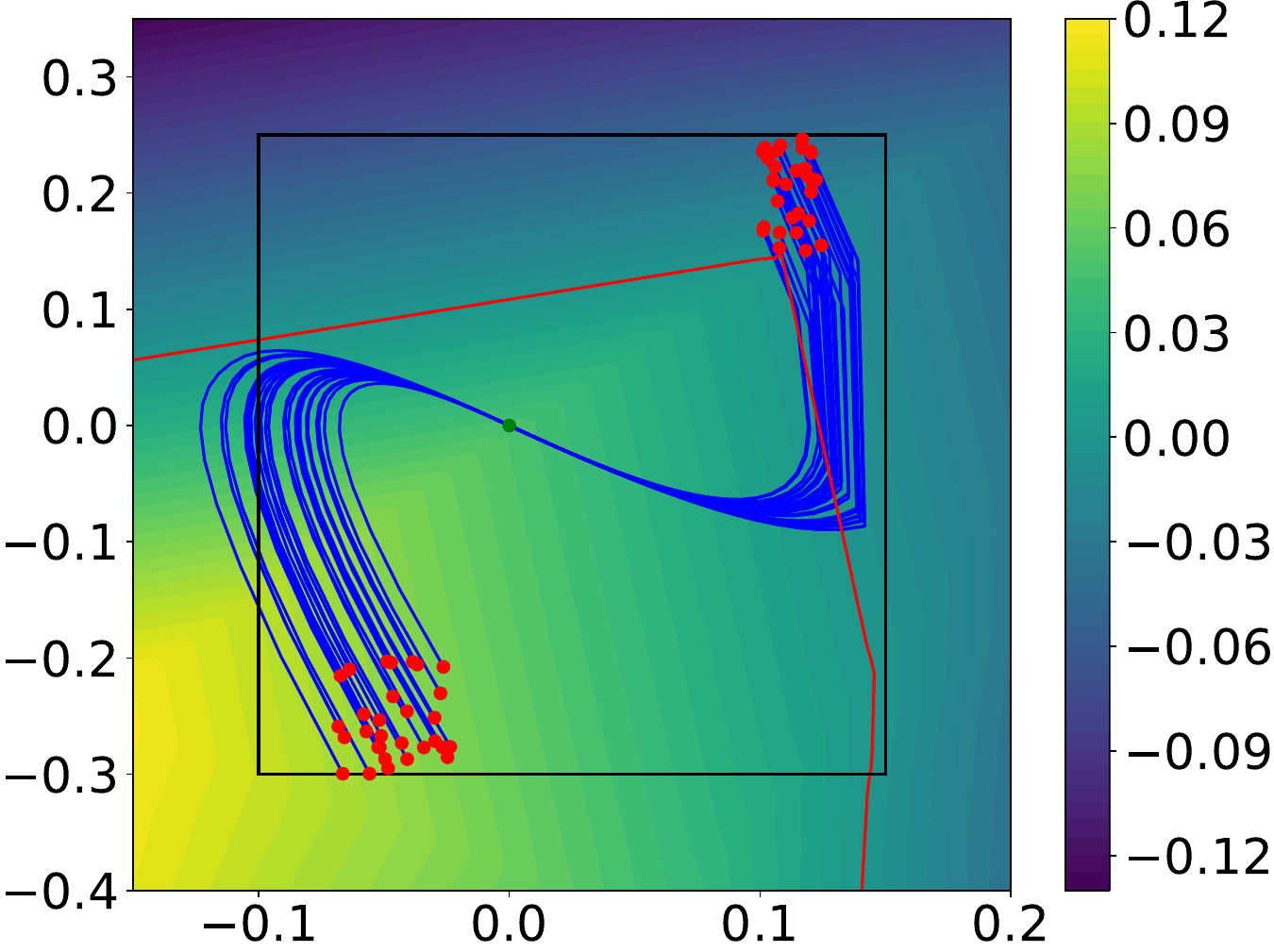}
    }
    \subcaptionbox{ROCBF\label{subfig:traj_rocbf}}{
    \includegraphics[width = .4\linewidth]{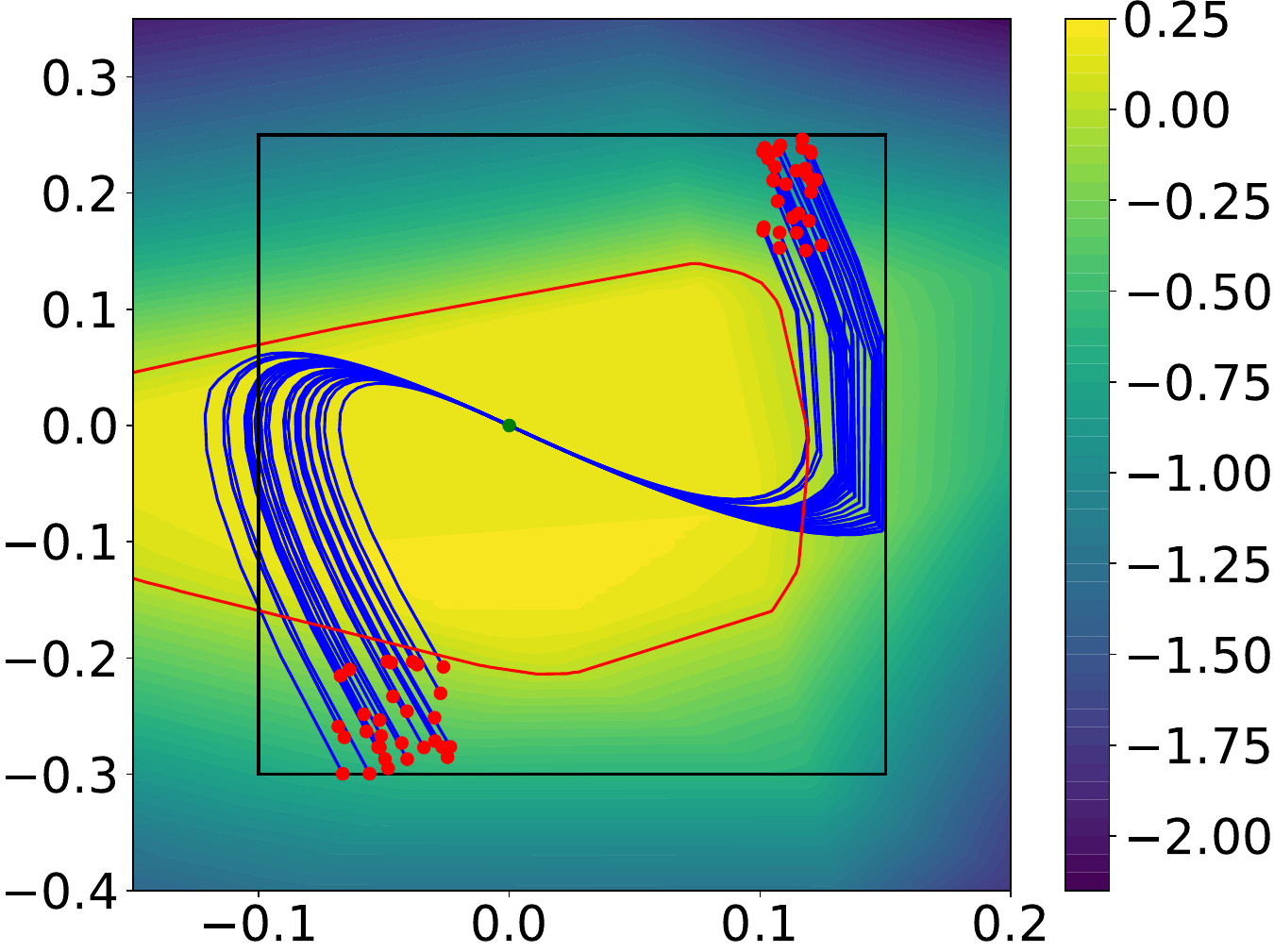}
    }
    \subcaptionbox{L-CBF\label{subfig:traj_gt}}{
    \includegraphics[width = .4\linewidth]{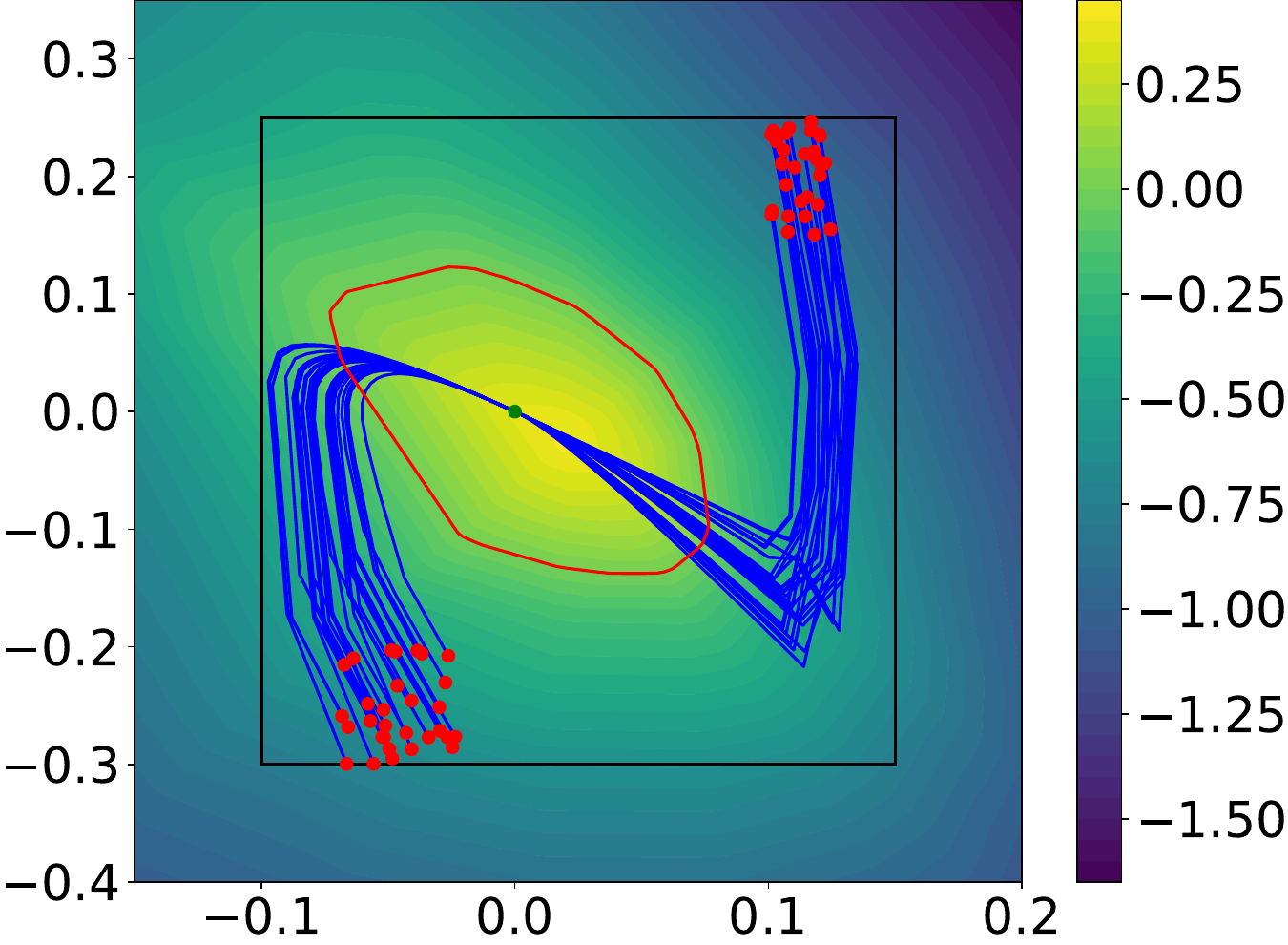}
    }
    % \hussein{the font of the numbers in the figures should be increased.}
    \caption{ The neural CBFs learned with different methods along with trajectories of the inverted pendulum  
    % using learned neural CBFs 
    starting from sampled initial states and following corresponding CBF-QP policies. 
    % that would result in unsafe trajectories if followed the reference controllers without safety filters 
    The red points are initial states, the green points (at the origin) are the goal states, and the red contours represent the zero level set of each neural CBF.}\label{fig:traj}
    \vspace{-0.5cm}
\end{figure}

Finally, we visualize a set of trajectories of the inverted pendulum generated using the different safety filters in  Figure \ref{fig:traj}.  
% We  analyze the performance of each method using trajectories sampled with $\pi_\mathit{ref}$ equipped with the learned CBFs for the . 
We sampled a set of initial states for which the reference controller would result in a collision with the absence of a safety filter. 
% The trajectories generated using ROCBF 
% suddenly diverge near the obstacle and that is likely because of ROCBF's misclassification of some safe states around the obstacle as unsafe. 
% negatively impacts SR due to its  misclassification of safe states around the origin as unsafe. 
The trajectories generated using iDBF and ROCBF reach the goal but enter the failure set. In contrast, both ICL-CBF and L-CBF were able to successfully intervene to prevent the system from entering the failure set and attract it towards their sublevel sets which contain the goal.
% while minimally affecting the progress towards reaching the goal.

% Thus, ICL-CBFs can effectively filter out unsafe actions without being overly conservative, thereby minimally affecting goal achievement.

\begin{table*}[h]
    \centering
    \scriptsize
    \caption{Success rates (SR) and collision rates (CR) of each method in different scenarios.}
    \begin{tabular}{ll|lll|l}
    \toprule
    Task & Metric & iDBF & ROCBF & ICL-CBF & L-CBF \\
    \midrule
    \multirow{2}{*}{Single integrator} & CR & $99.20\pm0.74$ & $\textbf{0.00}\pm\textbf{0.00}$ & $\textbf{0.00}\pm\textbf{0.00}$ & $0.00\pm0.00$ \\
    & SR & $0.80\pm0.74$  & $9.80\pm 2.14$ & $\textbf{80.60}\pm \textbf{4.41}$ & $86.20\pm 1.94$\\
    \midrule
    \multirow{2}{*}{Inverted pendulum} & CR & $2.80\pm1.17$ & $5.00\pm1.90$ & $\textbf{0.20}\pm\textbf{0.40}$ & $0.60\pm0.80$\\
    & SR & $97.20\pm1.17$ & $95.00\pm1.90$ & $\textbf{99.80}\pm\textbf{0.40}$ & $99.40\pm0.80$\\
    \midrule
    \multirow{2}{*}{Dubins car} & CR & $0.00\pm0.00$ & $69.30\pm23.30$ & $\textbf{1.80}\pm\textbf{1.33}$ & $0.30\pm0.64$\\
    & SR & $75.00\pm25.00$ & $6.40\pm3.44$ & $\textbf{97.60}\pm\textbf{1.50}$ & $99.60\pm0.80$\\
    \midrule
    \multirow{2}{*}{Quadrotor} & CR & $65.7\pm22.14$ & $75.00\pm25.00$ & $\textbf{17.10}\pm\textbf{6.64}$ & $1.50\pm1.12$\\
    & SR & $2.8\pm0.98$ & $0.00\pm0.00$ & $\textbf{77.20}\pm\textbf{4.31}$ & $98.00\pm1.26$\\
    \bottomrule
    \end{tabular}
    \label{tab:performance}
\end{table*}

\subsection{RQ2: Does the heuristic algorithm for approximating $\hat{\pi}_E$ degrade the quality of $B_\theta$?}
% \subsection{RQ2: Does the heuristic algorithm for approximating $\hat{\pi}_E$ degrade the performance of $B_\theta$?}
% Training a new CBF in each iteration can be time-consuming. To address this issue, we propose a heuristic version of our method. To answer the question outlined in RQ2, 
We evaluated the performance of ICL-CBF on the single integrator and quadrotor scenarios with and without  using the heuristic for approximating $\hat{\pi}_E$ during training.
% and compare the results with those when not non-heuristic settings. 
As shown in Table \ref{tab:efficiency}, training the ICL-CBF is significantly 
% more than two times 
slower when not 
% using  the original algorithm than when 
using the heuristic. The heuristic results in more noticeable degradation of CR and SR in the quadrotor scenario than the single integrator one. This is likely due to the higher dimensional state space of the quadrotor that requires finer resolution when choosing controls, and that is achieved using quadratic programming better than a brute-force search over coarse grids. The degradation might be an acceptable cost for the gains in training time for small enough dimensional systems. As the required grid size increases, the heuristic becomes slower and less useful. 
%it would become less advantageous to use it instead of the original algorithm. 
%one. 
% This training time gap would only increase with higher dimensions.
% settings.
% where sample complexity and computational time for retraining the CBF in every iteration would increase. 
% This difference will increase in higher-dimensional environments. 
% Both the heuristic and the original methods generate ICL-CBFs that result in zero CR. 
% The ICL-CBF generated using the original algorithm achieves a slightly higher SR.
% This demonstrates that the heuristic method efficiently trains ICL-CBFs without significant effects on performance. 
% with minimal effects on their performance.
% achieving performance comparable to the original approach.

% \yuxuan{
% Since the estimated $\hat{\pi}_E$ already provides a controller that adheres to $\pi_\mathit{ref}$ while satisfying the learned constraint $\hat{c}_\phi$, one might question the necessity of retraining a CBF for safe control. 
% To demonstrate the advantages of ICL-CBF, 
% Of course, the improvement in training time obtained from using the heuristic depends on the inference time of the grid-based policy. The latter in turn depends on the  size of the grid, which depends on the dimension and the volume of $\mathcal{U}$. As the grid size increases, the heuristic becomes slower and it would become less advantageous to use it instead of the original algorithm. 

Moreover, we compared the inference times of the generated CBF-QP policies with those of grid-based policies. They follow the same logic as the ones obtained using the heuristic. They use the learned constraint directly without training the ICL-CBF in the last iteration. We tried using grids for inference with two different resolutions. The first is the same as the one used for training the constraint and the second is a larger one.  
% after convergence along with grids of different resolutions over the control space.
% $\hat{\pi}_E$ when it is the CBF-QP based on ICL-CBF
% , i.e., the current output of Algorithm~\ref{alg:icl} while using the heuristic, 
% and when it is the grid-based policy, obtained at the last iteration without training the ICL-CBF.
% , in the single integrator and quadrotor scenarios.  
% Our aim is to check if there is any benefit in training a neural CBF using the learned constraint when one can use the latter directly to obtain safe control as done in the first $N-1$ iterations.  
% The latter is the one obtained by following the heuristic in the last iteration instead of training a neural CBF. 
% Thus, it uses the learned constraint function to generate the safe policy directly without training a neural CBF. 
%To compare their inference time, 
We generated five trajectories per scenario, each with 1000 time steps using both policies and computed the average trajectory-wise total inference time. The results are shown also in Table~\ref{tab:efficiency}. They show that although using the grid-based policy with a small grid size might decrease the inference time, it often comes at the cost of worse performance, while an ICL-CBF and its corresponding CBF-QP policy significantly  outperform them while having small enough inference time, especially in higher dimensions.  
\begin{table*}[h]
    \centering
    \scriptsize
    \caption{Success rates (SR) and collision rates (CR), training time, and inference time of different methods in the single integrator and quadrotor scenarios.}
    \begin{tabular}{lllllll}
    \toprule
    Task & Method & CR & SR & Training Time & Inference Time\\
    \midrule
    \multirow{4}{*}{Single integrator} 
    & ICL-CBF-based $\hat{\pi}_E$  (without heuristic) & $\textbf{0.00}\pm\textbf{0.00}$ & $\textbf{81.80}\pm\textbf{2.32}$ & $578.11\pm61.52$ & $3.67\pm0.18$ \\
    & ICL-CBF-based $\hat{\pi}_E$ (with heuristic) & $\textbf{0.00}\pm\textbf{0.00}$ & $80.60\pm4.41$ & $217.80\pm2.40$ & $3.77\pm 0.14$\\
    & Grid-based $\hat{\pi}_E$ (grid size of $50^2$) & $\textbf{0.00}\pm\textbf{0.00}$ & $79.40\pm3.67$ & $\textbf{189.80}\pm\textbf{2.42}$ & $\textbf{0.83}\pm\textbf{0.09}$\\
    & Grid-based $\hat{\pi}_E$ (grid size of $250^2$)& $\textbf{0.00}\pm\textbf{0.00}$ & $80.40\pm6.18$ & $\textbf{189.80}\pm\textbf{2.42}$ & $4.52\pm0.23$\\
    \midrule
    \multirow{3}{*}{Quadrotor} & ICL-CBF-based $\hat{\pi}_E$ (without heuristic) & $\textbf{9.30}\pm\textbf{4.12}$ & $\textbf{87.60}\pm\textbf{3.44}$ & $2967.33\pm148.97$ & $4.15\pm 0.29$\\
    & ICL-CBF-based $\hat{\pi}_E$  (with heuristic) & $17.10\pm6.64$ & $77.20\pm4.31$ & $2369.38\pm 270.49$ & $4.27\pm0.39$\\
    &Grid-based $\hat{\pi}_E$ (grid size of $100^2$) & $32.10\pm 11.18$ & $57.20\pm 4.12$ & $\textbf{2318.28}\pm\textbf{271.26}$ & $\textbf{3.52}\pm\textbf{0.12}$  \\
    &Grid-based $\hat{\pi}_E$ (grid size of $300^2$)& $24.20\pm9.00$ & $67.60\pm4.96$ & $\textbf{2318.28}\pm\textbf{271.26}$ & $29.70\pm4.11$\\
    \bottomrule
    \end{tabular}
    \label{tab:efficiency}
    % \vspace{-0.3cm}
\end{table*}

\subsection{RQ3: Should we generate $\mathcal{X}_S^B$ using $\pi_\mathit{ref}$ or $\hat{\pi}_E$?}
% To ensure the performance of the trained model $B_\theta$, the set $\mathcal{X}_S$ must contain a sufficient number of unsafe states, particularly those that the expert avoids while following the reference controller. Typically, trajectories generated by the safety-agnostic reference controller will include such unsafe states, otherwise a safety filter would not be needed. Hence, 
% our method 
Algorithm~\ref{alg:icl} constructs $\mathcal{X}_S^B$ using  $\pi_\mathit{ref}$, resulting in safe and unsafe trajectories. Recall that if the heuristic for generating $\hat{\pi}_E$ is followed, $\mathcal{X}_S^B$ is only generated  once in the $N^{\mathit{th}}$ iteration. If it were to be generated using the heuristic-based $\hat{\pi}_E$ of the $(N-1)^{\mathit{th}}$ iteration, it will likely contain few unsafe states as $\hat{c}_\phi$ would have converged, and that would not result in a balanced $\mathcal{X}_S^B$ for training $B_\theta$.   
Alternatively, one can construct $\mathcal{X}_S^B$ in the $N^{\mathit{th}}$ iteration as the union of all $\mathcal{X}_S^c$ over all iterations, i.e.,  by aggregating all the trajectories sampled using $\hat{\pi}_E$ in all  $N-1$ iterations. 
% Algorithm~\ref{alg:icl} would classify their trajectories using $\hat{c}_\phi$ in line~\ref{ln:x_safe}. 
% , as CBF-QP might still lead the agent into states the expert avoids before $\hat{c}_\phi$ converges.
We compare this strategy with the one using $\pi_{\mathit{ref}}$ in the single integrator scenario.
The results, shown in Table \ref{tab:data}, demonstrate that the $B_\theta$ learned from data sampled with $\pi_\mathit{ref}$ outperforms the one trained with the union of the $\mathcal{X}_S^c$ sets in terms of both CR and SR. We hypothesize that the reason is that while training  $\hat{c}_\phi$, it is not sufficiently accurate. That results  in the distribution of states in the union of  the $\mathcal{X}_S^c$ sets  different from the one encountered during deployment. 

\begin{table}[h]
 \vspace{-0.1in}
    \scriptsize
    \centering
    \caption{Success rates (SR) and collision rates (CR) of neural CBFs trained using data sampled using different policies in the  single integrator scenario.}
    \begin{tabular}{lllllll}
    \toprule
    Data Source & Collision Rate & Success Rate\\
    \midrule
    Sampled using $\pi_\mathit{ref}$& $\textbf{0.00}\pm\textbf{0.00}$ & $\textbf{80.60}\pm\textbf{4.41}$ \\
    Union of $\mathcal{X}_S^c$'s & $2.20\pm1.17$ & $78.60\pm3.50$ \\
    \bottomrule
    \end{tabular}
    \label{tab:data}
    % \vspace{-0.5cm}
\end{table}
%Some trajectories sampled with $\hat{c}_\phi$ might deviate significantly from the expert set, thus becoming noise in the data.

\subsection{RQ4: Are the safety labels produced using the learned constraint   accurate?}\label{sec:data}
\begin{figure}
    \centering
    \subcaptionbox{ICL-CBF\label{subfig:data_icl}}{
    \includegraphics[width = .4\linewidth]{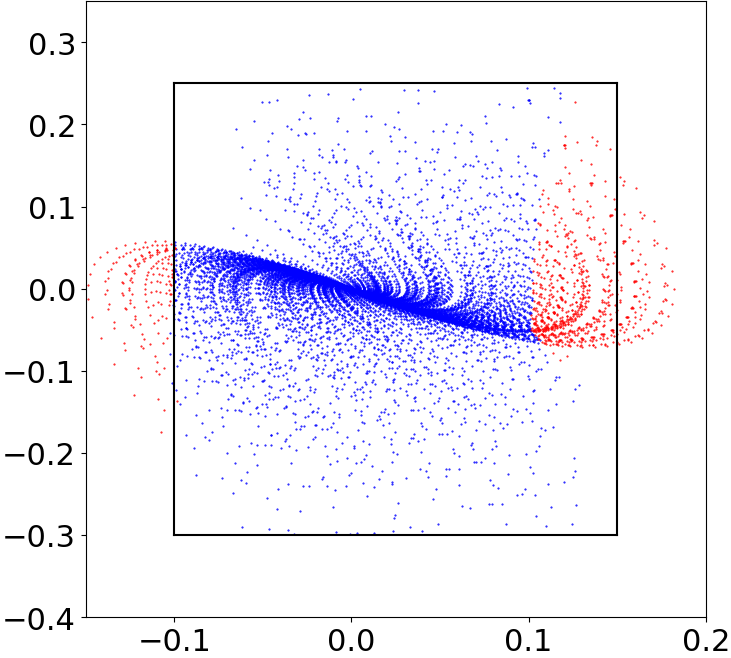}
    }
    \subcaptionbox{iDBF\label{subfig:data_idbf}}{
    \includegraphics[width = .4\linewidth]{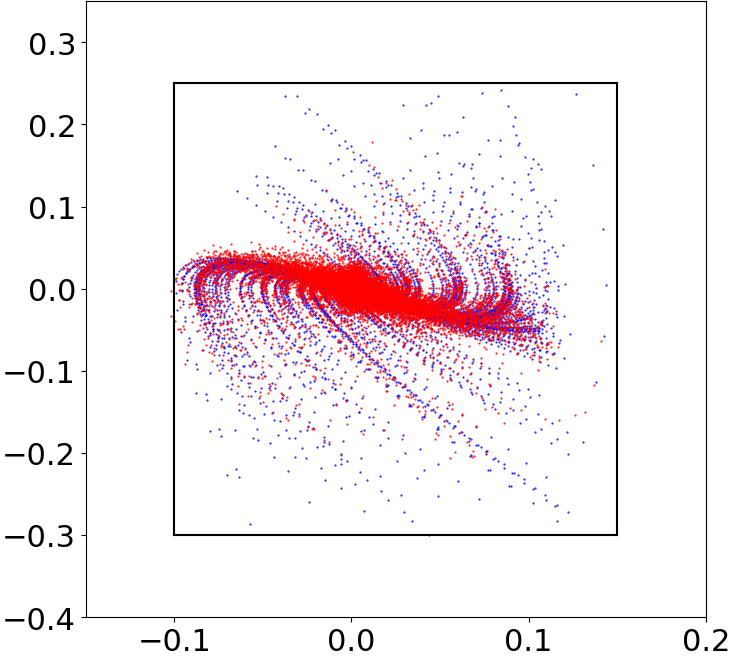}
    }
    \subcaptionbox{ROCBF\label{subfig:data_rocbf}}{
    \includegraphics[width = .4\linewidth]{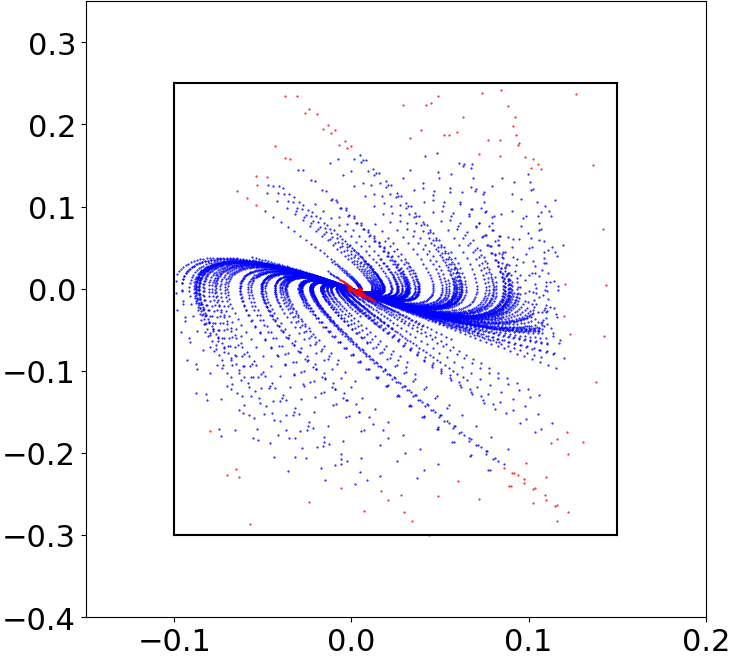}
    }
    \subcaptionbox{L-CBF \label{subfig:data_gt}}{
    \includegraphics[width = .4\linewidth]{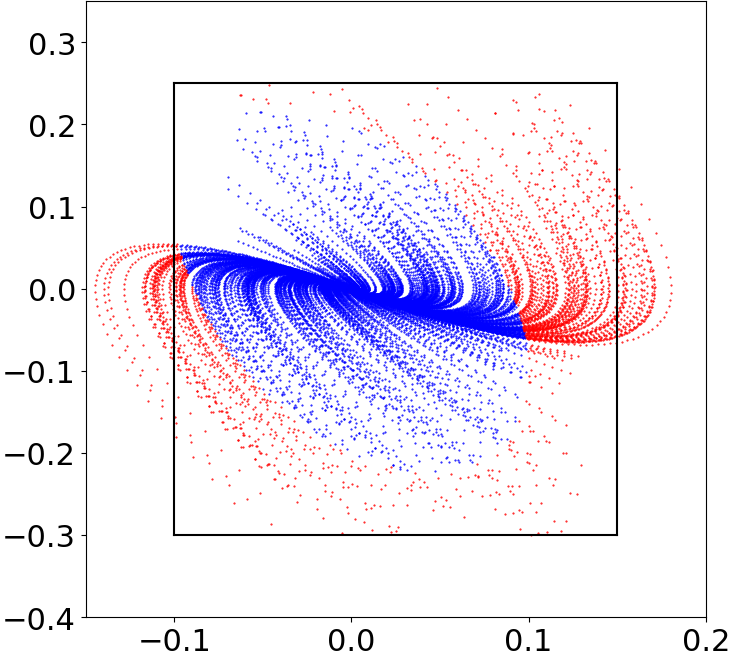}
    }
    \caption{Training data generated with different methods for inverted pendulum, where states within the black margins belong to the predefined safe set. Red points represent unsafe states, while blue points indicate safe states as annotated by each method.}\label{fig:data}
    \vspace{-0.4cm}
\end{figure}

We subjectively analyze the accuracy of the labels used in the  training of different neural CBFs in the inverted pendulum scenario. As discussed earlier, each training method labels the states visited by the expert trajectories and sampled ones, if any, differently. 
% \hussein{I'll wait for the response to my question on slack regarding the figure shown in case the colors are based on the learned CBF or the logic used during training for labeling sampled trajectories. } 
% Here, we use the training data of L-CBF as the upper bound since it's labeled by the ground truth CBF. 
The results are shown in Figure \ref{fig:data}, where the red points represent  states labeled as unsafe and blue points represent  states labeled as safe. The labels of the states generated by iDBF in this low-dimensional state space are inaccurate because the  states labeled as unsafe are generated by applying actions to each expert state. When 
% the time step of each state is small and 
the expert states are close to each other, the  states sampled and labeled as unsafe are close to the expert safe states. This issue should be a less of a concern in high-dimensional state spaces. As for the labels generated by ROCBF, only states at the boundary of the region where the expert trajectories reside are classified as unsafe, avoiding having overlapping states with different labels. However, states near the origin were considered boundary points by the reverse-KNN algorithm which lead ROCBF’s method to mislabel them as unsafe. 
% accurately samples only when states are evenly distributed. 
The constraint learned by MT-ICL, however, accurately classifies most states inside the failure set  as well the states starting from which lead to failure states as unsafe. Its results are the closest to the true labels used by L-CBF shown in Figure~\ref{subfig:data_gt}. % Although some states labeled as unsafe in L-CBF are marked as safe in ICL-CBF, the majority of these states are indeed safe, as they reside within the safe region and are unlikely to transition into the unsafe region subsequently. 
% Based on these results, we demonstrate that the data annotations generated by ICL-CBF are accurate.

\begin{figure}
    \centering
    \includegraphics[width=1\linewidth]{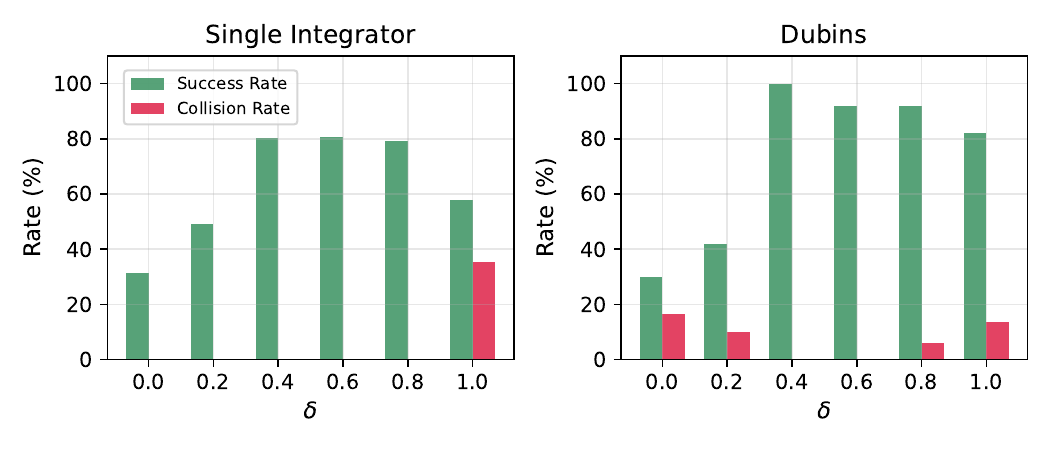}
    \caption{\yuxuan{Collision and success rates as $\delta$ varies in the Single integrator and Dubins car scenarios.}}
    \label{fig:sensitivity}
    \vspace{-0.4cm}
\end{figure}

\subsection{RQ5: How sensitive is the performance of the learned ICL-CBF  to the choice of the hyperparameter $\delta$?}
%, and can $\delta$ be selected efficiently in practice?}
\yuxuan{In our algorithm, $\delta$ serves as the threshold we use to partition the set $\mathcal{X}^B_S$. Because the constraint is trained on data produced by this partition, its performance is highly sensitive to the choice of $\delta$. 
%We therefore conduct a sensitivity analysis. 
As shown in Figure~\ref{fig:sensitivity}, the performance of ICL-CBFs can deteriorate when $\delta$ is poorly chosen. Nevertheless, we found searching for a $\delta\in[0,1]$ results in one with reasonable performance. In practice, one can discretize the interval $[0,1]$ and perform a grid search to find the best one when evaluated at the validation data. In our experiments, $\delta=0.6$ consistently produced strong results, although further tuning may yield additional gains.}

\section{CONCLUSION}\label{sec:conclusion}
% In this paper, we propose a method that trains CBFs with data labeled by the classifier learned from a variation of an ICL algorithm. 
In this paper, we address the problem of training neural CBFs from safe expert demonstrations using inverse constraint learning. 
% We propose a method for training neural CBFs using trajectories labeled by a classifier trained  using an ICL algorithm. 
We use ICL to learn a constraint function and then use it to label the states in newly sampled trajectories as safe or unsafe. Our approach requires a set of expert demonstrations, a potentially unsafe task-achieving reference controller, and the system dynamics. We compare our method against two baseline algorithms as well as against neural CBFs trained using labeled data. Our empirical results validate the effectiveness of our method in generating neural CBFs that improve safety while minimally affecting performance.

% \subsection{Training details}
% We train two-layer MLPs as the constraint models for the single integrator and inverted pendulum, and three-layer MLPs as the constraint models for the \yuxuan{Dubins} car and quadrotor scenarios. We train three-layer MLPs as the neural CBFs for all the tasks. Other hyperparameters used in our paper can be found in Table~\ref{tab:hyperparams}. We report the average performance along with the standard deviation over five runs for each trained model.

% \begin{table}[h]
%     \centering
%     \caption{Hyperparameters used in our paper.}
%     \begin{tabular}{lllllll}
%     \toprule
%      & Single integrator & Inverted pendulum & \yuxuan{Dubins} car & Quadrotor\\
%     \midrule
%     $\hat{c}_\phi$ Architecture & $[2,32,1]$ & $[2,32,1]$ & $[3,64,64,1]$ & $[6,128,128,1]$\\
%     $B_\theta$ Architecture & $[2,32,16,1]$ & $[2,32,16,1]$ & $[3,64,64,1]$ & $[6,128,128,1]$\\
%     $\hat{c}_\phi$ Learning Rate & $3e^{-4}$ & $3e^{-4}$ & $3e^{-4}$ & $3e^{-4}$\\
%     $B_\theta$ Learning Rate & $3e^{-5}$ & $3e^{-5}$ & $3e^{-5}$ & $3e^{-5}$\\
%     $\epsilon_\mathit{safe}$ & 0.2 & 0.2 & 0.2 & 0.2\\
%     $\epsilon_\mathit{unsafe}$ & 0.2 & 0.3 & 0.2 & 0.2\\
%     $\epsilon_\mathit{ascent}$ & 0.05 & 0.05 & 0.05 & 0.05\\
%     $w_\mathit{safe}$ & 1 & 1 & 1 & 1\\
%     $w_\mathit{unsafe}$ & 1 & 1.5 & 1 & 1.5\\
%     $w_\mathit{ascent}$ & 1 & 1 & 1 & 1\\
%     $\delta$ & 0.6 & 0.3 & 0.4 & 0.6\\
    
%     \bottomrule
%     \end{tabular}
%     \label{tab:hyperparams}
% \end{table}

\bibliographystyle{IEEEtran}
\bibliography{Yuxuan}
\end{document}

%% file: newcommands.tex
\usepackage{graphicx}
\AtBeginDocument{%
  \providecommand\BibTeX{{%
    \normalfont B\kern-0.5em{\scshape i\kern-0.25em b}\kern-0.8em\TeX}}}

\usepackage{graphicx}
\usepackage{wrapfig}
\usepackage{booktabs} % For formal tables

\usepackage{url}
\usepackage{amssymb,amsmath}
\usepackage{subcaption}
\usepackage{graphicx}
\usepackage{comment}

\usepackage{amsthm}
\usepackage{thmtools, thm-restate}
\usepackage{multirow}

\usepackage[graphicx]{realboxes}

\usepackage{avant} % Use the Avantgarde font for headings
\usepackage{mathptmx} % Use the Adobe Times Roman as the default text font together with math symbols from the Sym­bol, Chancery and Com­puter Modern fonts

\usepackage{microtype} % Slightly tweak font spacing for aesthetics
\usepackage[utf8x]{inputenc} % Required for including letters with accents
\usepackage{tikz}
\usetikzlibrary{shapes}
\RequirePackage{ifthen}
\RequirePackage{amsmath}
\RequirePackage{amsthm} %looks much better with this
\RequirePackage{amssymb}
\RequirePackage{xspace}
\RequirePackage{graphics}
\usepackage{keyval}
\usepackage{xspace}
\usepackage{mathrsfs}

\DeclareMathAlphabet{\mathcal}{OMS}{cmsy}{m}{n}

\newcommand{\yuxuan}[1]{\textcolor{black}{#1}}

\newcommand{\dom}{\relax\ifmmode {\mathit{dom}} \else ${\sf dom}$\fi}

\usepackage{setspace}
\usepackage{listings}
\newcounter{theorems}
\newtheorem{theorem}[theorems]{Theorem}%[theorem]

 \newcounter{remarks}
 
\newtheorem{remark}[remarks]{Remark}%[remark]

\newcounter{definitions}
 \newtheorem{definition}[definitions]{Definition}%[definition]

\usepackage[normalem]{ulem}

\usepackage{placeins}
\makeatletter
\newcommand\nocaption{%
    \renewcommand\p@subfigure{}
    \renewcommand\thesubfigure{\thefigure\alph{subfigure}}
}
\makeatother